\documentclass[lettersize,journal]{IEEEtran}
\usepackage{amsmath,amsfonts}
\usepackage{algorithmic}
\usepackage{algorithm}
\usepackage{array}
\usepackage[caption=false,font=normalsize,labelfont=sf,textfont=sf]{subfig}
\usepackage{textcomp}
\usepackage{stfloats}
\usepackage{url}
\usepackage{verbatim}
\usepackage{graphicx}
\usepackage{cite}
\usepackage{multirow}
\usepackage{bm}
\usepackage{booktabs}
\usepackage[colorlinks,linkcolor=blue]{hyperref}

\begin{document}

\title{HPPS: A Hierarchical Progressive Perception System for Luggage Trolley Detection and Localization at Airports}

\author{Zhirui Sun, Zhe Zhang, Jieting Zhao, Hanjing Ye, and Jiankun Wang, \emph{Senior Member, IEEE} % <-this % stops a space
\thanks{This work is supported by National Natural Science Foundation of China under Grant 62103181 and Shenzhen Science and Technology Program under Grant 20231115141459001. \emph{(Corresponding author: Jiankun Wang).}}% <-this % stops a space
\thanks{Zhirui Sun, Zhe Zhang and Jiankun Wang are with Shenzhen Key Laboratory of Robotics Perception and Intelligence, Department of Electronic and Electrical Engineering, Southern University of Science and Technology, Shenzhen, China (e-mail: sunzr2023@mail.sustech.edu.cn; 12232106@mail.sustech.edu.cn; wangjk@sustech.edu.cn).}%
\thanks{Zhirui Sun and Jiankun Wang are also with Jiaxing Research Institute, Southern University of Science and Technology, Jiaxing, China.}%
\thanks{Jieting Zhao and Hanjing Ye are with Shenzhen Key Laboratory of Robotics and Computer Vision, Department of Electronic and Electrical Engineering, Southern University of Science and Technology, Shenzhen, China (e-mail: jietingz2000@gmail.com; teamedlar@gmail.com).}%
}



\maketitle

\begin{abstract} 
The robotic autonomous luggage trolley collection system employs robots to gather and transport scattered luggage trolleys at airports. However, existing methods for detecting and locating these luggage trolleys often fail when they are not fully visible. To address this, we introduce the Hierarchical Progressive Perception System (HPPS), which enhances the detection and localization of luggage trolleys under partial occlusion. The HPPS processes the luggage trolley’s position ($x,y$) and orientation ($\theta$) separately, which requires only RGB images for labeling and training, eliminating the need for 3D coordinates and alignment. The HPPS can accurately determine the position of the luggage trolley with just one well-detected keypoint and estimate the luggage trolley's orientation when it is partially occluded. Once the luggage trolley's initial pose is detected, HPPS updates this information continuously to refine its accuracy until the robot begins grasping. The experiments on detection and localization demonstrate that HPPS is more reliable under partial occlusion compared to existing methods. Its effectiveness and robustness have also been confirmed through practical tests in actual luggage trolley collection tasks. A website about this work is available at \href{https://sites.google.com/view/robot-hpps}{HPPS}.
\end{abstract}

% \def\abstractname{Note to Practitioners}
% \begin{abstract}
% The motivation of this article is to detect and locate the luggage trolleys even when they are partially occluded. The effective perception of luggage trolleys is the basis for autonomous luggage trolley collection tasks, but the existing methods cannot work well under occlusion. Then, we propose the HPPS, which integrates the hierarchical process structure and progressive perception strategy. The innovative hierarchical structure separately identifies the position and orientation of the luggage trolley. The progressive perception continuously updates the luggage trolley's pose during the navigation. The HPPS works well even when the luggage trolleys are partially occluded, ensuring they are detected and located reliably. Additionally, the HPPS only needs RGB images for labeling and training, enhancing the scalability and practical implementation. The effectiveness and practicality of HPPS have promoted the implementation of the robotic autonomous luggage trolley collection system at airports.
% % which efficiently perceives the position and orientation of luggage trolleys, even in cases of partial occlusion.
% \end{abstract}


\section{Introduction}
\IEEEPARstart{I}{ntelligent} robotics and autonomous driving technologies \cite{trajectory-prediction} \cite{intelligent-vehicles} are increasingly utilized in areas like traffic lighting \cite{ts}, logistics \cite{logistics-2}, and mining trucks \cite{mt}, playing a significant role in their advancement. Similarly, researchers are exploring the robotic system for autonomous luggage trolley collection at airports. It means using robots to gather luggage trolleys scattered around the airport and transport them to designated areas for reuse. It helps reduce the need for human resources and improve efficiency in collecting luggage trolleys. Designing such a robotic system is complex and involves multiple components, such as object detection, localization, motion planning, control, and manipulation. 
% \IEEEPARstart{A}{s} intelligent robotics and autonomous driving technologies advance \cite{trajectory-prediction} \cite{intelligent-vehicles}, their applications are expanding to enhance efficiency in areas like traffic lighting \cite{ts}, underwater vehicle \cite{uv}, and mining trucks \cite{mt}. Similarly, researchers are exploring the robotic system for autonomous luggage trolley collection at airports. 
\begin{figure}[t]
    \centering
    \includegraphics[width=1\columnwidth]{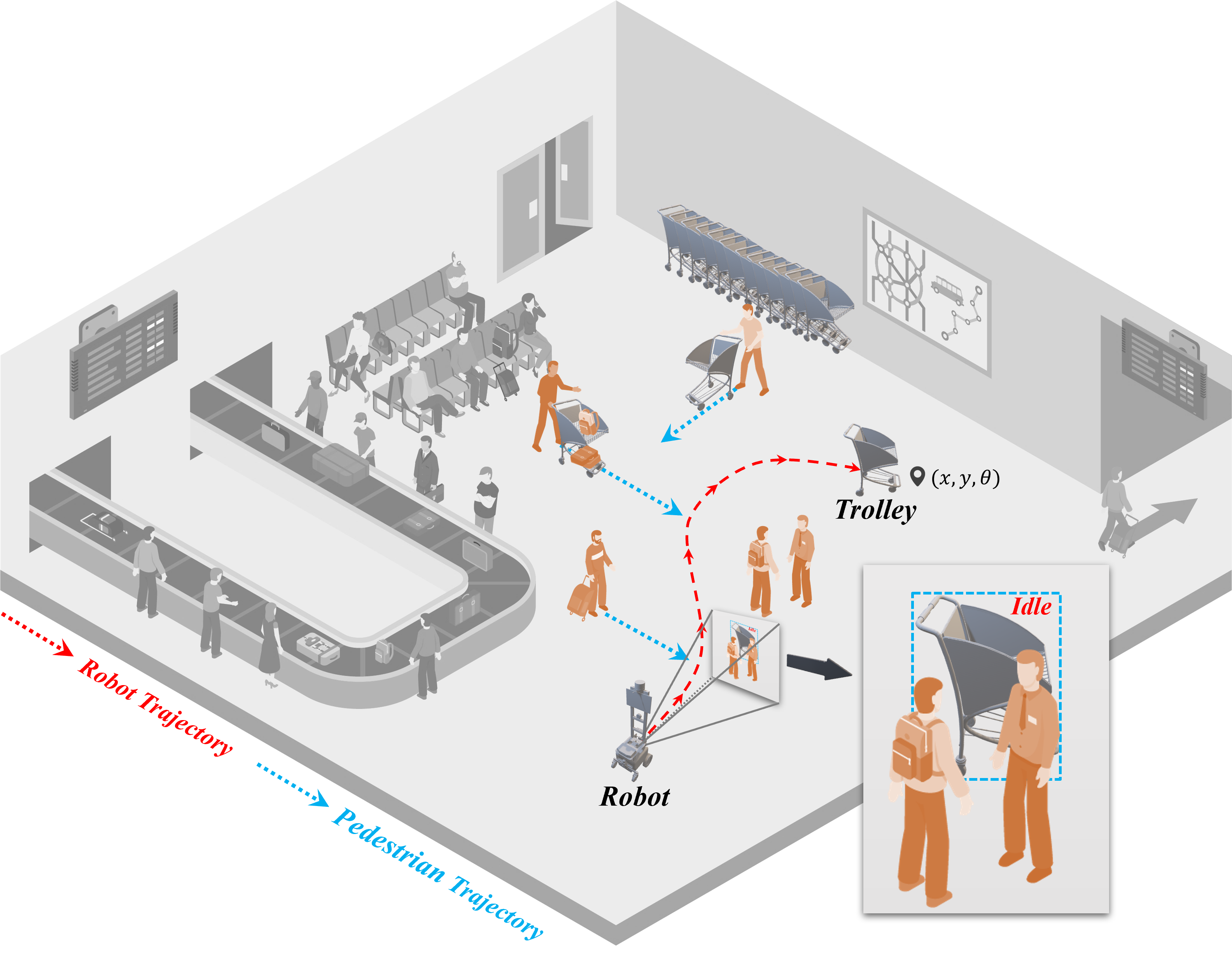}
    \caption{Schematic diagram of HPPS for luggage trolley detection and localization under partial occlusion at airports.}
    \label{top}
\end{figure}
Some progress has been made in areas like robotic autonomous trolley collection \cite{xiao}, autonomous multiple-trolley collection \cite{xie} and collaborative trolley transportation \cite{xby}. However, these studies usually ignore situations where the luggage trolley is occluded, or do not consider the perception of the luggage trolley. Challenges remain in detecting and localizing luggage trolleys, especially in occlusion cases. The luggage trolley often becomes partially occluded due to objects or people at airports, as illustrated in Fig. \ref{top}, where two people stop to talk, occluding the view of the idle luggage trolley. Therefore, effective detection and localization under partial occlusion are essential for the system. They ensure the robot to identify the correct luggage trolley's pose, which is fundamental for subsequent motion planning and accurate manipulation of end-effectors.

For detecting luggage trolleys, current methods overlook scenarios where the luggage trolley is occluded and fail to distinguish the usage states of the luggage trolley. It leads to frequent failures in detecting luggage trolleys and the inability to identify idle ones from various usage states. To address this, a new dataset of luggage trolleys is collected and labeled. This dataset includes different luggage trolley states, multiple occlusion levels, and various environmental settings. Retraining with this dataset effectively resolves the issue of failing to detect idle luggage trolleys under partial occlusion. For luggage trolley localization, mainstream methods rely on wireless, visual, and laser sensors. Choosing the most suitable method based on the application scenario presents a challenge. Fortunately, previous studies have already demonstrated solutions to this. In related work \cite{tim}, multiple localization methods are systematically evaluated. It concludes that vision-based Keypoints methods are most suitable for robotic autonomous luggage trolley collection system. It primarily relies on gathering keypoint information from the luggage trolley and determining the luggage trolley's pose by solving EPNP \cite{epnp}. However, when the luggage trolley is occluded or when keypoint detection is incomplete, this method often fails to determine the luggage trolley's pose.

% In this work, we propose a hierarchical progressive perception system that can realize the detection and localization of luggage trolleys under partial occlusion. % This suitability is influenced by many factors like localization accuracy, mobile power supplies, coverage area, and scalability, among others. However, this method still has the potential for further improvement.
In this work, we propose HPPS, which detects and locates luggage trolleys even under partial occlusion. Traditional methods for acquiring the pose of a luggage trolley face significant limitations, particularly in handling occlusions and ensuring accuracy and real-time. For example, the method by Xiao \emph{et al}. \cite{xiao}, which relies on solving EPNP, fails to operate effectively under occluded conditions. The approach by Pan \emph{et al}. \cite{pan}, which uses the registration of point clouds, struggles with poor real-time performance and insufficient pose accuracy. HPPS processes and obtains the position and orientation of the luggage trolley independently, which can handle complex situations efficiently and robustly. HPPS requires only simple RGB information to determine the luggage trolley's position and orientation without complex sensor equipment. For input RGB images, HPPS identifies the luggage trolley's keypoints via a detection network and then calculates the luggage trolley's position using the camera's projection geometry relationship. At the same time, HPPS's orientation detection network predicts the luggage trolley's orientation probability distribution and then regresses the most likely orientation. The luggage trolley's pose is determined by integrating these two parts of information. In this process, we divide the pose of the luggage trolley into position and orientation, processing them separately and simplifying the requirements for input information. Compared to methods requiring depth cameras or LiDAR, this system only needs a monocular camera, making it more cost-effective. RGB image data is more accessible than depth data, allowing the use of many existing image datasets and network resources, thus reducing the workload of data collection and preprocessing. Moreover, once the initial pose of the luggage trolley is obtained, the robot continues to detect the luggage trolley during navigation to refine the localization accuracy further until the grasping process begins.
\subsection{Contributions}
The main contributions of this article are summarized as follows:
\begin{itemize}
    \item This article presents a dataset of 13740 images capturing various luggage trolley states and occlusion levels across diverse environments to improve luggage trolley detection accuracy in real-world scenarios.
    \item This article introduces a novel Hierarchical Progressive Perception System (HPPS) designed for detecting and locating luggage trolleys under partial occlusion.
    \item Real-world experiments demonstrate the robustness and accuracy of HPPS in complex and dynamic environments, where it detects and locates the target luggage trolley and successfully collects it. This progress enhances the deployment of robotic autonomous luggage trolley collection systems at airports.
\end{itemize}
\subsection{Outline}
The remainder of this article is organized as follows. The related work is reviewed in Sec. \ref{related-work}. Sec. \ref{HPPS} provides the details of the HPPS, including the Detection Module, Keypoints Process Module, Orientation Process Module, Filter Module, and Motion Planner Module. The Dataset, Implementation Details, Experiment Platform, and Results are explained in Sec. \ref{experiments}. The final section, in Sec. \ref{future}, summarizes this article and considerations for future work.

\section{Related Work}\label{related-work}
\subsection{2D Bounding-box Detection and Classification}
Object detection and classification are essential for identifying and localizing objects within images. Popular methods include region-based approaches like the R-CNN series \cite{R-CNN}\cite{Fast-R-CNN}\cite{Faster-R-CNN} and single-shot detectors such as YOLO (You Only Look Once) \cite{yolo-review} and SSD (Single Shot MultiBox Detector) \cite{ssd}.
While R-CNN models are highly accurate, they are computationally intensive. In contrast, YOLO offers a compelling balance between speed and accuracy, making it suitable for real-time applications. YOLO divides the image into a grid, each cell predicting multiple bounding boxes and their confidence scores, which indicate the likelihood of an object's presence and the accuracy of the box. Each grid cell also classifies the detected object, combining localization with precise categorization.
Considering both performance and efficiency, the YOLO model is an ideal choice for real-time object detection. Its ability to process images quickly and with high accuracy meets the demands of most practical applications. Employing the YOLO model will facilitate efficient and accurate visual recognition capabilities for our task.

\subsection{Keypoint Detection}
Keypoint detection is instrumental for accurately identifying and localizing specific parts of objects in images. Several neural network models have significantly advanced this field:
OpenPose \cite{openpose} excels in multi-object scenarios by generating part candidates and analyzing their connections.
Stacked Hourglass Networks \cite{hourglass} leverage a repetitive structure that captures and integrates features at multiple scales, ideal for precise single-object keypoint localization.
DeepPose \cite{deeppose}, developed by Google, applies deep learning to keypoint detection by directly regressing coordinates from images but may struggle with complex backgrounds.
Convolutional Pose Machines (CPM) \cite{cpm} refine keypoint predictions through multi-stage processing, each enhancing the heatmaps' accuracy.
HRNet \cite{hrnet} maintains high-resolution representations throughout the network, enabling simultaneous capture of detailed and contextual information, making it highly effective for complex poses and environments.
HRNet maintains high-resolution streams throughout its architecture, effectively capturing fine and coarse features. This design allows HRNet to excel in precision and detail, outperforming other models, especially on challenging datasets like COCO keypoints. Unlike other models that may lose spatial information due to pooling operations, HRNet retains high-resolution features during processing, making it particularly suitable for complex scenarios involving occlusions and various poses.

\subsection{Orientation Estimation}
Accurate orientation estimation is fundamental for autonomous systems. Traditional approaches relied on handcrafted features and machine learning classifiers like Support Vector Machines (SVMs) \cite{svm}, which often struggled with occlusions and complex backgrounds. Current deep learning-based methods frequently approach orientation estimation through classification tasks. The method in \cite{cnn} employs a four-layer neural network to regress the orientation directly from the image, providing an end-to-end solution. In the study by \cite{cnn+}, the orientation is classified into eight bins and then regressed within those bins for a finer estimate. These methods employ straightforward network structures, leading to models that achieve optimal performance in environments similar to the training data used. Yu \emph{et al}. \cite{usingkeypoints} present models that detect keypoints and infer orientation based on the spatial arrangement of these keypoints, effectively handling occlusions. Estimating orientation directly from images is valid, as it simplifies the annotation process for training datasets and enhances performance by concentrating on the orientation estimation task \cite{prove}.
Considering the strengths and constraints of these methods, combining HRNet and ResNet provides a promising solution. HRNet excels in keypoint detection, and ResNet is suited for global feature extraction and orientation estimation. This combined method offers reliable orientation estimation, maintaining precision even when objects are partially occluded.
% Combining HRNet's keypoint detection prowess with ResNet's orientation regression capacity can create a robust orientation estimation system capable of handling complex occlusions while delivering accurate orientation information.
% Considering the strengths and constraints of these approaches, merging HRNet's keypoint detection with ResNet's feature extraction offers a compelling strategy. HRNet is proficient at detecting keypoints, and ResNet is apt for global feature analysis and inferring orientation. This integrated approach promises a resilient system for orientation estimation, adept at managing occlusions and ensuring precise orientation data.

\subsection{3D Pose Estimation}
3D pose estimation is crucial for determining objects' spatial orientation and position. Current leading methods in this domain primarily utilize deep learning. These approaches employ neural networks to determine the pose of objects from the input images. 
PoseCNN \cite{posecnn} predicts the 3D rotation and translation directly from images, suitable for single-object scenarios but struggles with complex backgrounds and occlusions. It also requires extensive, precisely annotated 3D pose data, which can be costly and time-consuming to prepare.
DeepIM \cite{deepim} improves initial pose estimates through an iterative process, offering significant precision advantages. However, its high computational cost and slow processing speed make it unsuitable for real-time applications. DeepIM also heavily relies on accurate initial pose estimations.
PVNet \cite{pvnet} uses keypoints and a voting mechanism to estimate poses, handling partial occlusions and various viewing angles effectively. However, it demands precisely annotated 2D image positions and corresponding 3D spatial locations, which can be challenging with data collection.
NOCS \cite{nocs} provides an end-to-end solution by mapping objects into a normalized coordinate space, facilitating viewpoint and size-independent representations. While NOCS excels at generalizing to unseen object categories, its reliance on detailed 3D models and precise data alignment during training limits its practical use.
MonoLoc \cite{monoloc} utilizes a neural network to identify keypoints within an image, then employs these calculated 2D keypoint positions to get the object's 3D location through a multi-task neural network. 
EPro-PnP \cite{Epro-pnp} infers the object's pose using a 3D bounding box incorporating learnable 2D-3D correspondences.
The training datasets for these methods often require complex data collection and annotation, which increases implementation costs. Furthermore, these datasets usually need complete observation of the objects. However, in environments like airports, luggage trolleys are frequently only partially visible, making these methods challenging to handle effectively.  

Compared to these methods, our approach separately detects the position and orientation of the luggage trolley. It uses a keypoints detector to provide 2D information for model-based 3D location processing and an orientation detector to estimate the orientation probability for Gaussian regression analysis. The main advantages of our approach include: 1) Utilizing 2D detectors simplifies the 3D pose estimation process by relying solely on RGB image data and offers improved robustness against partial occlusion compared to deep learning-based methods. 2) Our model-based process module efficiently uses a predefined model of the luggage trolley to compute position from well-detected 2D keypoints. 3) Our method decreases the requirement for dataset preparation, streamlines implementation, and enhances processing speed, making it well-suited for real-time applications, especially in robotic systems that are resource-constrained and time-sensitive.

\section{System Description}\label{HPPS}
\begin{figure*}[htb]
    \centering
    \includegraphics[width=2\columnwidth]{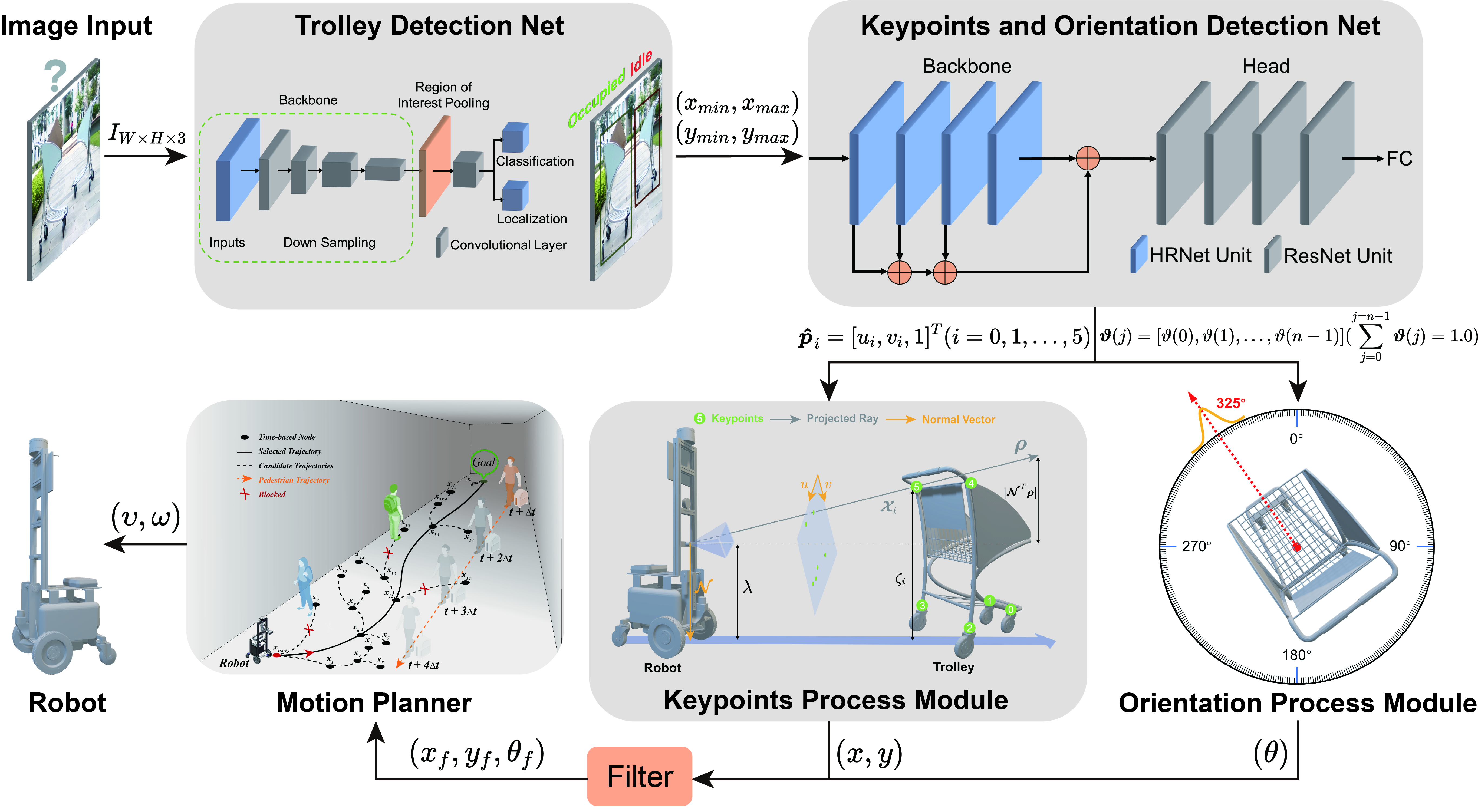}
    \caption{A diagram of our proposed luggage trolley detection and localization system. The process starts with capturing RGB images, followed by a detection phase. Next, the system processes these detection information to determine the luggage trolley's pose, which is then filtered and progressively updated. Finally, the planning module computes and sends velocity commands to the robot in real time.}
    \label{framework}
\end{figure*}
This section introduces the architecture of the luggage trolley detection and localization system, as shown in Fig. \ref{framework}. This system utilizes a camera to capture RGB images. The luggage trolley detection network identifies the idle ones within these images, marking them with bounding boxes. Based on these bounding boxes, the keypoints detection network gets the keypoints' coordinates on the image plane, and the orientation detection network predicts the luggage trolley orientation's probability distribution (Sec. \ref{detection}). The 3D coordinates of keypoints are determined through the camera's projection geometry and a prior model (Sec. \ref{keypoints}). The luggage trolley's orientation is estimated via Gaussian regression (Sec. \ref{Orientation}). Following a filtering process (Sec. \ref{filter}), the pose of the luggage trolley is continuously updated through effective observed information, thereby enhancing accuracy. After obtaining the luggage trolley's pose, this information guides the motion planning module (Sec. \ref{motion planning}) to generate control instructions for robot navigation. In the forthcoming definitions, boldfaced variables signify vectors, whereas non-bold variables represent scalars.

\subsection{Detection Module}\label{detection}
Given an input image $I$ with dimensions $W \times H \times 3$, the idle luggage trolley is identified, generating bounding boxes $[x_{min},y_{min},x_{max},y_{max}]$. From these bounding boxes, we can derive 2D coordinates of the keypoints on the image plane and a probability distribution of the luggage trolley's orientation. Specifically, YOLOV5 \cite{yolov5} is employed for real-time luggage trolley detection due to its effectiveness in object detection. The next step involves cropping the image within the bounding box to focus solely on the luggage trolley. This produces a cropped image $I_c\in \mathbb{R}^{W_c \times H_c \times 3}$. 

Inspiration by human pose estimation, we use HRNet to predict heatmaps of six 2D keypoint (as shown in Fig. \ref{keypoints-process}) coordinates $\boldsymbol{p}_i = [x_i, y_i]^T(i = 0,1,...,5)$, leading to the generation of homogeneous 2D keypoints coordinates $\boldsymbol{\hat{p}}_i = [u_i, v_i, 1]^T(i = 0,1,...,5)$. The HRNet and ResNet are combined \cite{hboe} for orientation detection. The cropped images pass through a backbone network, acting as a feature extractor. These features are then combined and processed via additional residual layers, culminating in a fully connected layer and a softmax layer. The outcome includes $n$ unit orientations, ${\vartheta}(j) = [\vartheta(0),\vartheta(1),...,\vartheta(n-1)](\sum_{j=0}^{j=n-1}{\vartheta}(j) = 1.0)$, each indicating the probability of corresponding orientation unit that best represents the luggage trolley's orientation in the image.

\subsection{Keypoints Process Module}\label{keypoints}
\begin{figure}[htb]
    \centering
    \includegraphics[width=1\columnwidth]{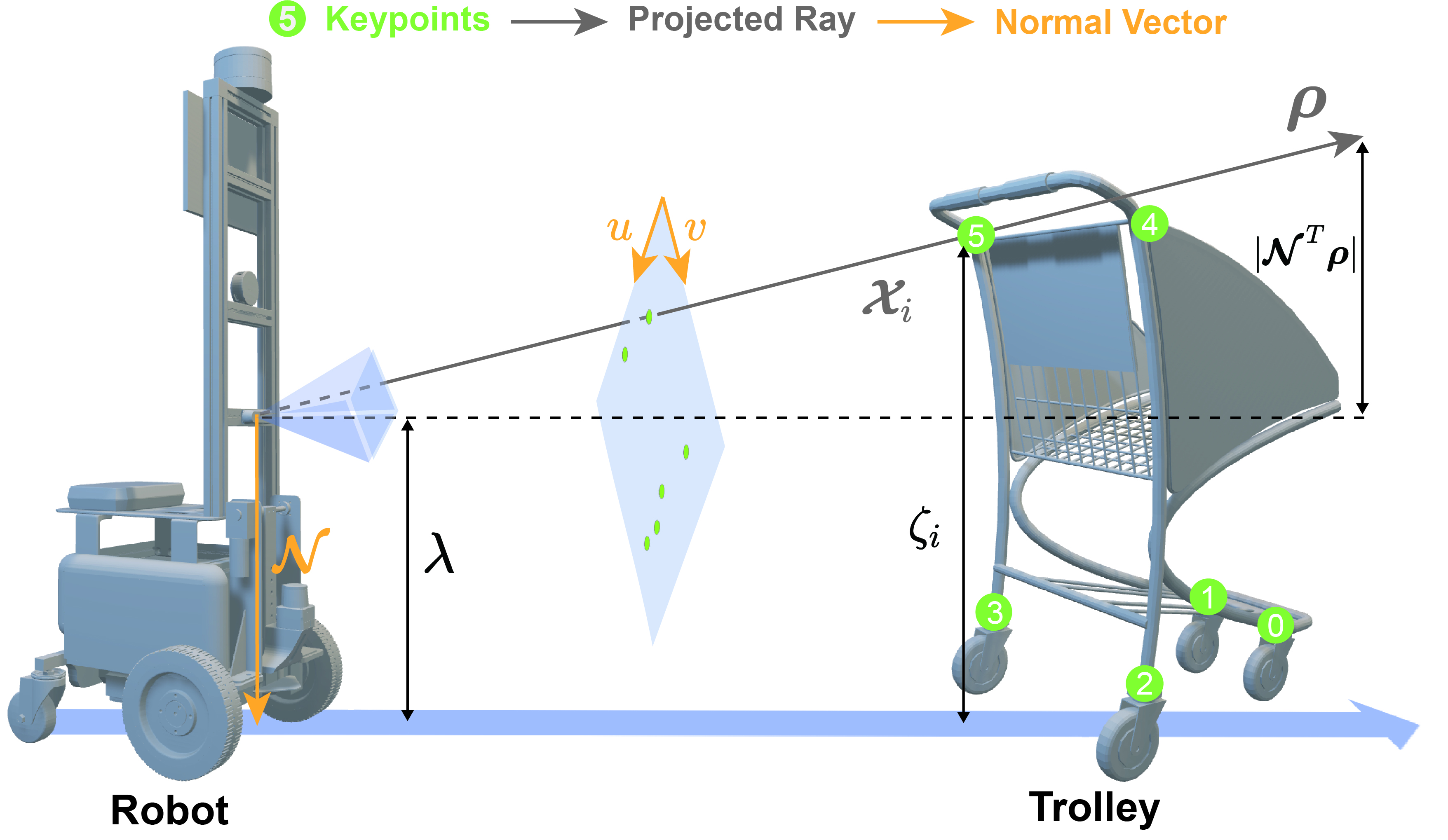}
    \caption{An example of estimating the luggage trolley's position through keypoint observation. The process utilizes any visible keypoint for the luggage trolley's position estimation. The parameters involved and the estimation method are detailed in Eq. \ref{estimation}.}
    \label{keypoints-process}
\end{figure}
The keypoints detection network identifies the 2D coordinates of the keypoints on the image plane. Assuming the center point of the luggage trolley is on the ground, a coordinate system is established with this center point as the origin. From this reference, the 3D coordinates of the keypoints are determined. With the known 3D heights of the keypoints corresponding to the center point, their precise 3D positions can be calculated \cite{yehanjing}. The luggage trolley's prior model is defined as follows:
\begin{equation}
\boldsymbol{\mathcal{M}} = \{\boldsymbol{\mathcal{Y}}_i = (x_i, y_i, \zeta_i) \in \mathbb{R}^3 \mid i = 0,1,\ldots,5\},
\end{equation}
where $\boldsymbol{\mathcal{Y}}_i$ represents the $i$-th keypoint's 3D position relative to the center point, consisting of coordinates $x_i, y_i$ and height $\zeta_i$. We define the position of the luggage trolley as the center point, $\boldsymbol{\mathcal{Y}} \in \mathbb{R}^3$, within the camera coordinate frame, satisfying the ground plane constraint \cite{constrain}:
% Then, all keypoints of the prior model are assumed to be on the ground, set at zero height. This establishes $\{\boldsymbol{\mathcal{\hat{M}}} = {\boldsymbol{\mathcal{\hat{Y}}}_i = (x_i, y_i, 0) \in \mathbb{R}^3 \mid i = 0,1,\ldots,5}\}$. As a result, all keypoints' 3D position $\boldsymbol{\mathcal{\hat{Y}}}$ satisfy the ground plane constraint \cite{constrain}:
\begin{equation}
\boldsymbol{\mathcal{-N}}^T \boldsymbol{\mathcal{\hat{Y}}} + \lambda = 0,
\end{equation}
where $\boldsymbol{\mathcal{N}}$ ($\boldsymbol{\mathcal{N}}\in \mathbb{R}^3$) indicates the normal vector to the ground, and $\lambda$ ($\lambda>0$) is the distance from the camera's optical center perpendicular to the ground, as shown in Fig. \ref{keypoints-process}. 
% These principles allow for the calculation of each keypoint's 3D position.
% Further, we define all 3D keypoints positions is based on the ground, which is set at a height of 0, that means $\boldsymbol{\mathcal{\hat{M}}} = \{\boldsymbol{\mathcal{\hat{X}}}_i = (x_i, y_i, 0) \in \mathbb{R}^3 \mid i = 0,1,\ldots,5\}$. Thus, all $\boldsymbol{\mathcal{\hat{X}}}$ need to satisfy the ground plane constraint:
% \begin{equation}
% \boldsymbol{\mathcal{-N}}^T \boldsymbol{\mathcal{\hat{X}}} + \lambda = 0,
% \end{equation}
% where $\boldsymbol{\mathcal{\hat{X}}}$ is the coordinate position of the keypoint, the height is 0, $\boldsymbol{\mathcal{N}}$ ($\boldsymbol{\mathcal{N}}\in \mathbb{R}^3$) is the normal vector perpendicular to the ground plane, and $\lambda$ ($\lambda>0$) is the distance from the camera optical center to the ground plane. Based on these definitions, the position of the key point in 3D space can be calculated.
 
Based on the prior model, the 3D position of visible keypoints in any input image is determined as follows:
% \begin{equation}
% \begin{aligned}
% r = K^{-1}p_j \\
% \mathbf{X_j} = \frac{|\gamma - h_j|}{|N^Tr|}\cdot r \\
% \mathbf{X}_i = \mathbf{X}_j - h_j \mathbf{N}
% \end{aligned}
% \end{equation}
\begin{equation}
\begin{aligned}
\boldsymbol{\rho} = \boldsymbol{\mathcal{K}}^{-1}\boldsymbol{\hat{p}}_i, \\
\boldsymbol{\mathcal{X}}_i = \frac{|\lambda - \zeta_i|}{|\boldsymbol{\mathcal{N}}^T\boldsymbol{\rho}|} \cdot \boldsymbol{\rho}. \\
\end{aligned}
\label{estimation}
\end{equation}

The image coordinates of the $i$-th keypoint can be obtained through the keypoints detection network, creating homogeneous coordinates $\boldsymbol{\hat{p}}_i$. By multiplying the homogeneous coordinates with the inverse of the camera's intrinsic matrix $\boldsymbol{\mathcal{K}}^{-1}$, getting the ray $\boldsymbol{\rho}$ that passes through the $i$-th keypoint. Using the geometric relation between the camera's height $\lambda$ and the height of the $i$-th keypoint $\zeta_i$, the 3D position of this keypoint $\boldsymbol{\mathcal{X}}_i$ is determined. Once the coordinates of $i$-th keypoint $\boldsymbol{\mathcal{X}}_i$ are calculated, and with the prior model $\boldsymbol{\mathcal{M}}$, the center point coordinates of the luggage trolley are derived using the following formula:
\begin{equation}
\boldsymbol{\mathcal{C}} = \frac{1}{N_{vis}}\sum_{i=1}^{N_{vis}}(\boldsymbol{\mathcal{X}}_{i,vis} - \boldsymbol{\mathcal{Y}}_{i,vis}),
\end{equation}
where $\boldsymbol{\mathcal{C}}$ represents the center point coordinates of the luggage trolley, $N_{vis}$ is the count of visible keypoints, $\boldsymbol{\mathcal{X}}_{i,vis}$ refers to the 3D position of the $i$-th visible keypoint and $\boldsymbol{\mathcal{Y}}_{i,vis}$ corresponds to the $i$-th visible keypoint in prior model. 

\subsection{Orientation Process Module}\label{Orientation}
\begin{figure}[htb]
    \centering
    \includegraphics[width=0.8\columnwidth]{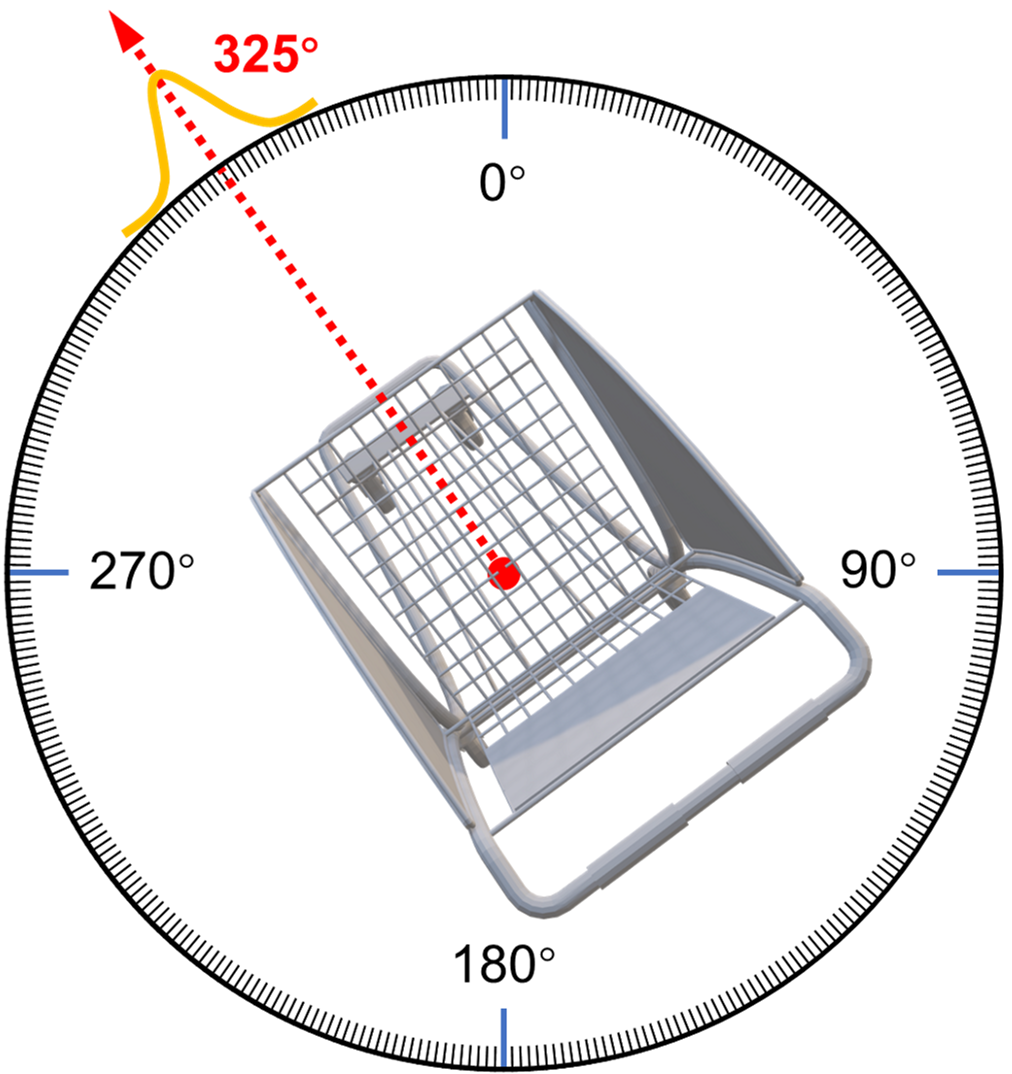}
    \caption{An example of estimating the luggage trolley's orientation from a top-down perspective. The yellow curve represents the Gaussian regression probability distribution of the orientation, while the red dotted line with an arrow indicates the orientation with the highest probability.}
    \label{orientation-process}
\end{figure}

The orientation detection network infers possible orientations for the luggage trolley. The number of possible orientations corresponds to the degree of discreteness. After conducting comparative experiments, this article adopts 360 divisions based on findings demonstrating optimal orientation accuracy. Details of the experimental results that support this decision are provided in Tab. \ref{train2}. ${\vartheta}(j)$ indicates the probability that the luggage trolley's orientation is within the $j$-th unit orientation. This means the orientation falls in the range of $\{\theta \mid j*1^\circ-0.5^\circ\leq \theta \leq j * 1^\circ+0.5^\circ\}$. The loss function for ${\vartheta}(j)$ is defined as follows:
\begin{equation}
\mathcal{J}(\theta)=\sum_{j=0}^{360}\left({\vartheta}(j)-\psi(\mu, \sigma)\right)^2,
\end{equation}
where $\psi(\mu, \sigma)$ represents the ``circular" Gaussian probability, a representation of the ground truth, shown in Fig. \ref{orientation-process} (orange curve):
\begin{equation}
\psi(\mu, \sigma)=\frac{1}{\sqrt{(2 \pi)} \sigma} e^{-\frac{1}{2 \sigma^2}\left(\min \left(\left|\mu-\tau\right|, 360-\left|\mu-\tau\right|\right)\right)^2},
\end{equation}
where $\tau$ is the ground truth orientation. This process predicts a Gaussian function centered on the accurate unit orientation. The idea is that a unit orientation closer to the true orientation gets a higher probability score from the model. Therefore, the final orientation, $\theta$, is determined by the highest probability score, represented as:
\begin{equation}
\theta = \arg\max({\vartheta}(j)) \times 1.
\end{equation}

% the expression for Q as follows:
% \begin{equation}
% Z = \sum_{k=0}^{71} \left(Q_k - \Theta(k, \beta)\right)^2,
% \end{equation}
% where Θ(k,β) symbolizes a "periodic" Gaussian function, which is exemplified in Fig. 3(b) (red line) by:
% \begin{equation}
% \Theta(k, \beta) = \frac{1}{\sqrt{2\pi} \beta} e^\left(-\frac{\left(\min\left(\left|k - l_{\text{opt}}\right|, 72 - \left|k - l_{\text{opt}}\right|\right)\right)^2}{2\beta^2}\right),
% \end{equation}
\subsection{Filter Module}\label{filter}
% Due to the occlusion challenge, the exact count of detectable keypoints is uncertain. 
To ensure the accuracy and stability of the luggage trolley's central coordinates, it is essential to implement filtering based on the coordinates of the visible keypoints. In this article, the Modified Moving Average Filter (MMAF) is employed, which dynamically adapts to incoming data points, focusing on mitigating the impact of outliers. The core process is described as follows:
\begin{enumerate}[]
\item Initialization of the Filter Parameters: 
\begin{equation}
\boldsymbol{\mathcal{F}} = (\Delta, \Theta_{z}),
\end{equation}
where $\Delta$ represents the window size for the moving average, and $\Theta_{z}$ denotes the threshold for z-score outlier detection.
\item Dynamic Adaptation and Outlier Filtering: Given a sequence of data points ${\mathcal{O}}_i = [x_i, y_i, \theta_i] \quad (i=1, 2, \ldots, N)$, the MMAF updates its state by: 
\begin{equation}
{\mathcal{Z}}({\mathcal{O}}_i) = \frac{|{\mathcal{O}}_i - {\mu}_{{\Delta}}|}{{\sigma}_{\Delta}},
\end{equation}
where ${\mu}_{\Delta}$ and ${\sigma}_{\Delta}$ denoting the mean and standard deviation of the points within the window $\Delta$, respectively. Computing the updated moving average excluding outliers:
\begin{equation}
\bar{{\mathcal{O}}} = \frac{1}{|N_{{\mathcal{Q}}}|} \sum_{{\mathcal{O}}_i \in {\mathcal{Q}}} {\mathcal{O}}_i, \forall {\mathcal{O}}_i \in {\mathcal{Q}}, {\mathcal{Z}}({\mathcal{O}}_i) \leq \Theta_{z},
\end{equation}
where ${\mathcal{Q}}$ is the set of data points not considered outliers, and $N_{{\mathcal{Q}}}$ represents the count of ${\mathcal{Q}}$.
\item Dynamic Adjustment:
\begin{equation}
\bar{{\mathcal{O}}}_f = 
\begin{cases} 
\bar{{\mathcal{O}}}, & \text{if } |{\mathcal{Q}}| > 0 \\
{\mathcal{O}}_{\text{latest}}, & \text{otherwise}
\end{cases}.
\end{equation}
This equation signifies the filter’s output is the average of non-outlier points if available; otherwise, it defaults to the latest point.
\end{enumerate}
\subsection{Motion Planner Module}\label{motion planning}
Using this filter, the luggage trolley's pose is progressively updated by effectively observing visible keypoints. Once the luggage trolley's pose is determined, the Motion Planner takes over the planning task. The Multi-Risk-RRT \cite{multi-risk-rrt}, our previous work, integrates Multi-directional Searching with Heuristic Sampling. This approach efficiently incorporates heuristic information from dynamic sub-trees into the rooted tree, overcoming the constraints of TBVP solvers. The Multi-Risk-RRT algorithm improves motion planning in static and dynamic environments, enabling the robot to receive control instructions based on the planned trajectory.
\begin{equation}
\begin{aligned}
\left[\upsilon, \omega\right] &= \text{Multi-Risk-RRT}\left(\left[x_f, y_f, \theta_f\right], Map\right),
\end{aligned}
\end{equation}
where $\upsilon, \omega \in \mathbb{R}^2$ are the linear and angular velocity of the robot, $x_f, y_f \in \mathbb{R}^2, \theta_f \in [0, 2\pi)$ are the pose of the luggage trolley, and $Map$ is the environment representation.
\section{Experiments}\label{experiments}
\begin{figure}[h]
    \centering
    \includegraphics[width=1\columnwidth]{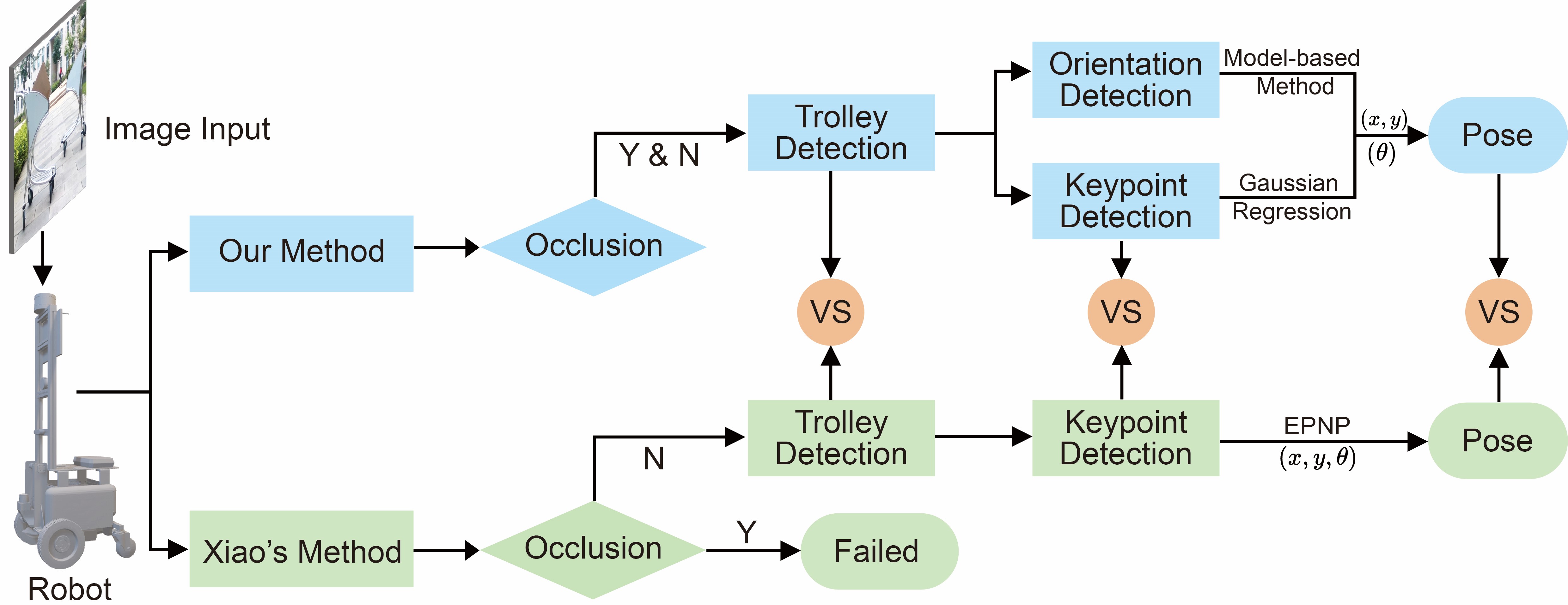}
    \caption{A workflow comparison of our method and Xiao's method for estimating the pose of the luggage trolley from image inputs.}
    \label{pipline}
\end{figure}
This section introduces the datasets, implementation details, experiment platform, and results. The experiments focus on detection, localization, and robot trials, illustrating the efficacy and resilience of our proposed HPPS. Fig. \ref{pipline} displays the experimental workflow, contrasting our method with Xiao's for estimating luggage trolley pose from image inputs. Initially, both methods assess their ability to detect luggage trolleys under occlusion. Our method successfully identifies luggage trolleys in such conditions, whereas Xiao's does not. Subsequently, our method employs separate processes for orientation and keypoint detection, contributing to the overall pose estimation through model-based methods and Gaussian regression. Conversely, Xiao's method applies an EPnP solver for pose estimation following keypoint detection. We select three critical stages during the experiment—luggage trolley detection, keypoint detection, and pose estimation—to compare the two methods.

\subsection{Dataset}
\begin{figure}[htb]
    \centering
    \includegraphics[width=1\columnwidth]{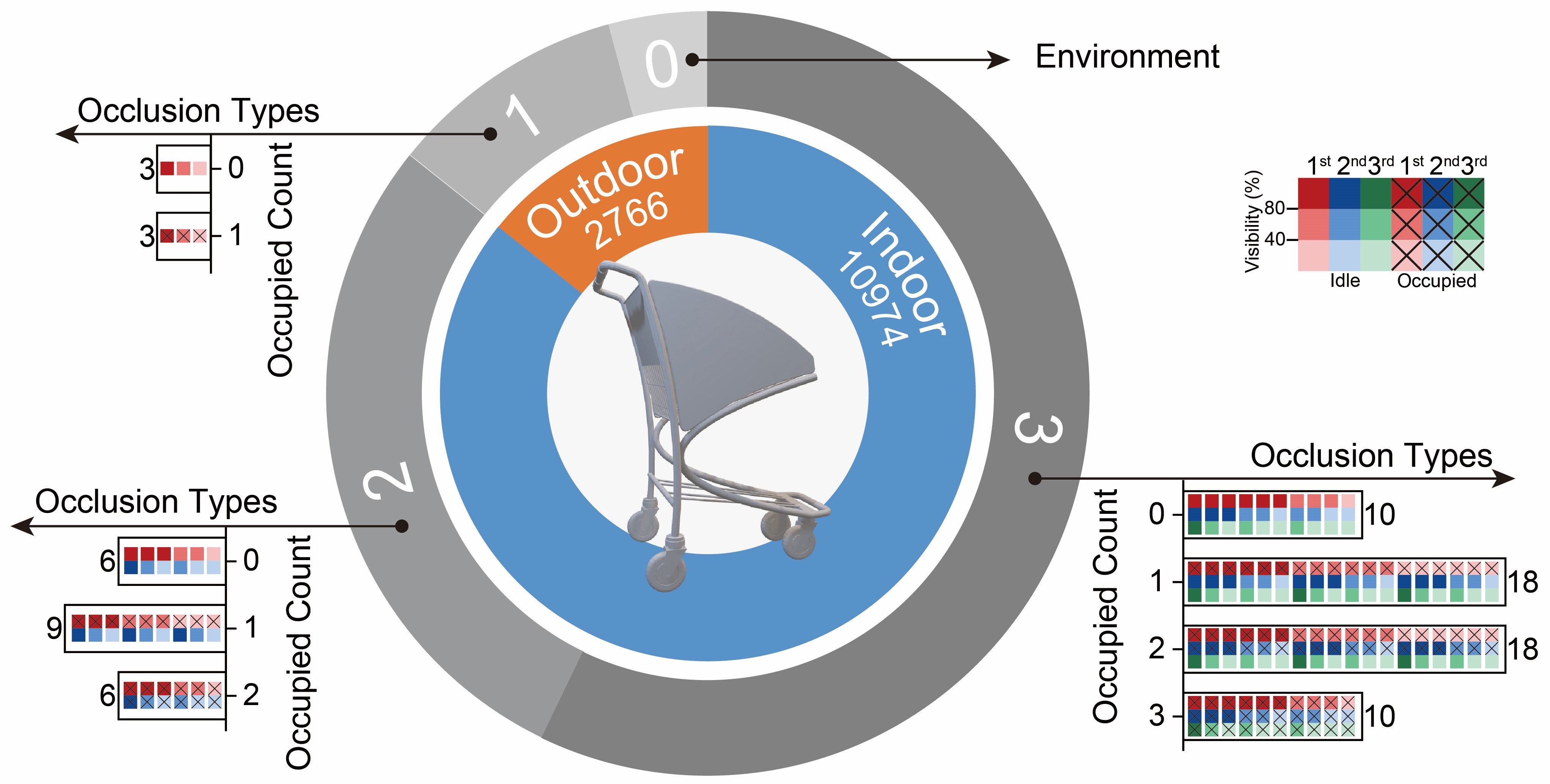}
    \caption{The diagram of dataset classification. The orange segment of the inner circle represents the images of the outdoor environment, and the blue segment represents the indoor environment. The outer circle's gradient gray illustrates the occupancy of luggage trolleys, ranging from 0 to 3. Each bar's height represents different types of occupied luggage trolleys, distinguished by red, blue, and green boxes. Transparency levels indicate visibility: high for under 40\%, medium for 40\%-80\%, and low for over 80\%. Crosses on the boxes mark occupied luggage trolleys.}
    \label{dataset}
\end{figure}

A detailed dataset of luggage trolleys is created to make the HPPS more accurate. The dataset comprises 13740 images featuring diverse backgrounds, lighting conditions, viewpoints, occlusion levels, and different luggage trolley states. Each image is labeled with 2D bounding boxes, six keypoints, an orientation angle, and an indicator to determine the usage states of the luggage trolley. This dataset is gathered in indoor and outdoor environments to enhance detection robustness. Recognizing that airports often have many luggage trolleys, this dataset includes images categorized by the number of luggage trolleys. To reflect real-world airport challenges, images capture luggage trolleys from various angles and include scenarios where luggage trolleys are either idle or occupied. As shown in Fig. \ref{dataset}, the inner circle highlights the contrast between indoor and outdoor environments, while the outer circle shows images based on the number of luggage trolleys. The label ``0" means no luggage trolley is in the image, and a bar chart shows the type distribution of images with occupied luggage trolleys under different occlusion conditions. The red, blue, and green boxes in each bar represent three different luggage trolleys. The transparency of each color indicates the luggage trolley's visibility level: high transparency shows visibility under 40\%, medium transparency denotes visibility between 40\% and 80\%, and low transparency indicates visibility over 80\%. Crosses in various colors on the boxes mark the occupied luggage trolleys. The dataset classifies the visibility and occupancy conditions of the luggage carts into 84 different categories.

\subsection{Implementation Details}
The dataset is divided into three parts: 80\% for training, 10\% for validation, and 10\% for testing. Training is performed offline using PyTorch on an AMD EPYC MILAN 7413 CPU and an NVIDIA RTX A6000 GPU. The 2D detection network builds on the official YOLOV5 \cite{yolov5} code. For training, the SGD \cite{sgd} optimizer assists for 300 epochs with a batch size 16. The training results are shown in Tab. \ref{train1}, which evaluates 1374 images across several metrics. 
% \textbf{Precision} calculates the fraction of correctly identified positive samples by the model among all positives predicted. \textbf{Recall} measures the fraction of actual positives accurately identified by the model out of all real positives. \textbf{mAP50} reflects the mean average precision at an IoU (Intersection over Union) threshold of 0.5. \textbf{mAP50-95} averages over IoU thresholds from 0.5 to 0.95, increasing in steps of 0.05. 
The results show the training model effectively identifies whether luggage trolleys are idle or occupied and accurately determines their bounding box.
\begin{table}[htb]\tiny
\centering
\caption{Training Results of the Luggage Trolley Detection and Classification}
\label{train1}
\resizebox{\columnwidth}{!}{%
\renewcommand{\arraystretch}{1.5}
\begin{tabular}{cccc}
\toprule[1pt]
\textbf{Metric} & \textbf{Overall} & \textbf{Occupied} & \textbf{Idle} \\ \hline
Images          & 1374             & 1374            & 1374          \\ 
Objects         & 3478             & 1706            & 1772          \\ 
Precision       & 0.973            & 0.97            & 0.977         \\ 
Recall          & 0.974            & 0.981           & 0.967         \\ 
mAP50           & 0.982            & 0.981           & 0.984         \\ 
mAP50-95        & 0.967            & 0.966           & 0.967         \\ \bottomrule [1pt]
\end{tabular}%
}
\end{table}

HRNet is utilized to detect 2D keypoints, and orientation detection is achieved through a combination of HRNet and ResNet. Training results ranging from 72 equal divisions to 360 equal divisions are analyzed to compare the relationship between angle discrete division and accuracy. Each training is conducted for 100 epochs with a batch size of 64. The training results are presented in Tab. \ref{train2}. Bins refer to equally spaced segments of the circle's 360 degrees. For example, dividing a circle into 72 equal parts means each bin corresponds to an angle of 5 degrees (360° / 72). Average degree error (ADE) indicates the average error in degrees. Acc.-5°, Acc.-15°, and Acc.-30° specify the rate at which predictions fall within the 5°, 15°, and 30° error margin, respectively. Keypoint detection accuracy (KDA) reflects the model's performance in identifying and accurately locating keypoints within images. It is evident that as the division becomes finer, the average error of the angle gradually decreases. From the comparison results, a division into 360 equal bins is selected, achieving an average degree error of less than 3° and a keypoint detection accuracy of 99.4\%.
% Tab. \ref{train2} displays the testing results of luggage trolley orientation errors and keypoint detection accuracy. Average degree error (ADE) indicates the average error in degrees. Acc.-5°, Acc.-15°, and Acc.-30° specify the rate at which predictions fall within the 5°, 15°, and 30° error margin, respectively. Keypoint detection accuracy (KDA) reflects the performance of the model in identifying and accurately positioning keypoints within images. Based on the test results, the average error in orientation detection is under 8 degrees, and the keypoint detection accuracy reaches 99.2\%.
% \usepackage{graphicx}

% \usepackage{graphicx}
% \begin{table}[htb]
% \centering
% \caption{Training Results of Orientation and Keypoints Detection}
% \label{train2}
% \resizebox{\columnwidth}{!}{%
% \renewcommand{\arraystretch}{1.5}
% \begin{tabular}{ccccl}
% \hline
% \textbf{ADE} & \textbf{Acc.-5°} & \textbf{Acc.-15°} & \textbf{Acc.-30°} & \textbf{KDA}                \\ \hline
% 7.54        & 70.20\%                    & 91.80\%                     & 97.90\%                     & \multicolumn{1}{c}{99.20\%} \\ \hline
% \end{tabular}%
% }
% \end{table}

\begin{table}[htb]
\centering
\caption{Comparison of Orientation and KeyPoint Detection Accuracy Across Various Angle Discretizations}
\label{train2}
\resizebox{\columnwidth}{!}{%
\renewcommand{\arraystretch}{1.5}
\begin{tabular}{cccccc}
\toprule[1pt]
\textbf{Bins} & \textbf{ADE}   & \textbf{Acc.-5°} & \textbf{Acc.-15°} & \textbf{Acc.-30°} & \textbf{KDA} \\ \hline
72                               & 9.005          & 62.60\%                    & 89.20\%                     & 97.20\%                     & 99.40\%      \\ 
90                               & 8.284          & 61.00\%                    & 88.40\%                     & 96.90\%                     & 99.40\%      \\ 
120                              & 6.909          & 61.20\%                    & 95.40\%                     & 97.50\%                     & 99.40\%      \\ 
180                              & 5.159          & 78.30\%                    & 97.00\%                     & 97.70\%                     & 99.40\%      \\ 
360                              & \textbf{2.572} & \textbf{95.20\%}           & \textbf{97.80\%}            & \textbf{98.60\%}            & 99.40\%      \\ \bottomrule [1pt]
\end{tabular}%
}
\end{table}

\begin{figure}[htb]
    \centering
    \includegraphics[width=1\columnwidth]{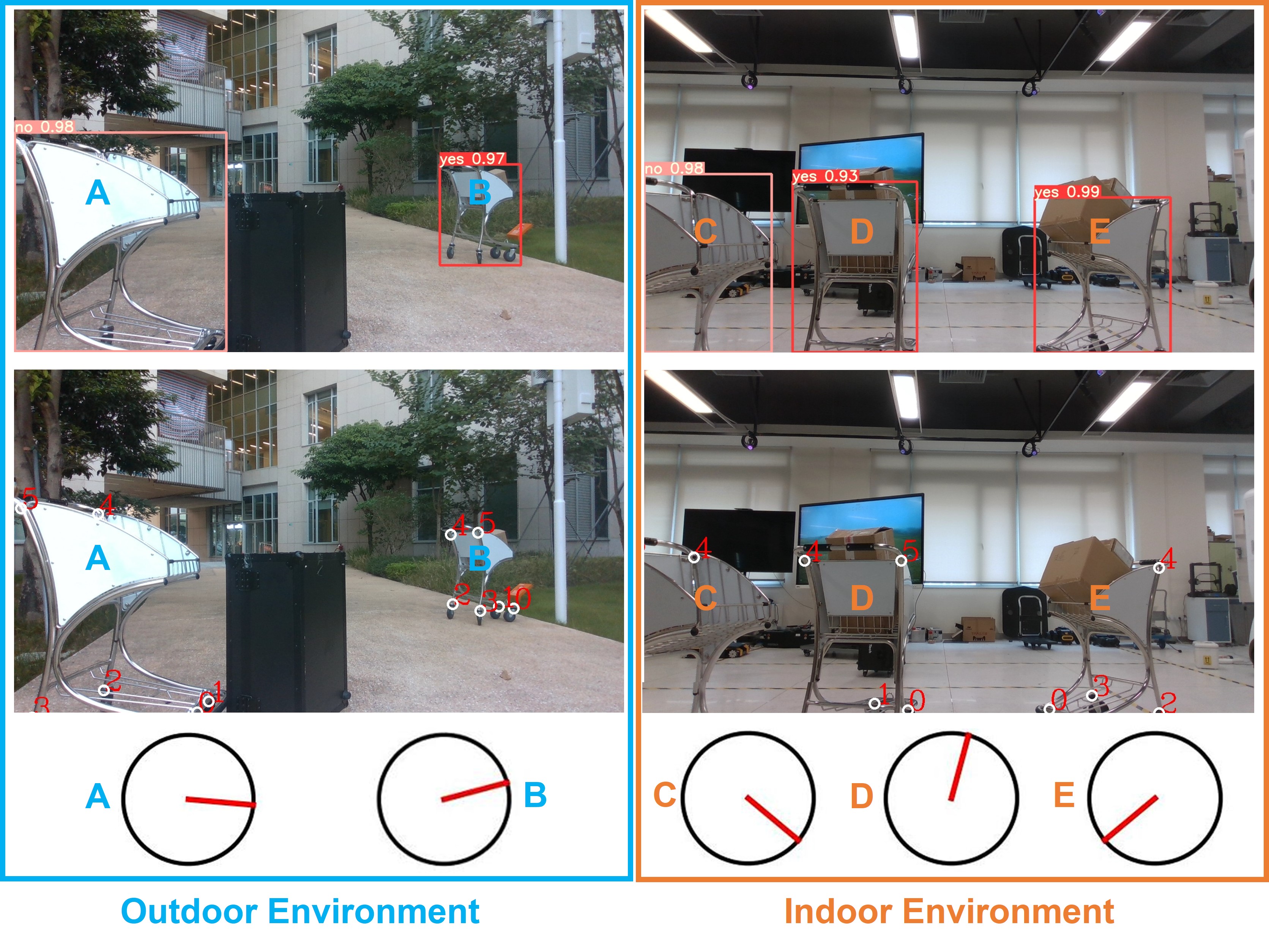}
    \caption{Visualization of detection results. The images are classified into two types of environments: outdoor and indoor. Consistent letters across the images denote various detection aspects of the same luggage trolley: the first row represents detection and classification, the second row shows keypoint detection, and the third row displays orientation prediction.}
    \label{testing-result}
\end{figure}
To enhance clarity, Fig. \ref{testing-result} displays two types of images: the left column for outdoor environments and the right column for indoor environments. The first row illustrates the detection and classification of luggage trolleys, with ``yes" indicating occupied and ``no" signifying idle. The second row highlights the detection of visible keypoints on the luggage trolleys, each marked with a white circle and a corresponding number. The third row displays the orientation of luggage trolleys, illustrated by a red line within a 360-degree circle.
\subsection{Experiment Platform}
\begin{figure}[htb]
    \centering
    \includegraphics[width=1\columnwidth]{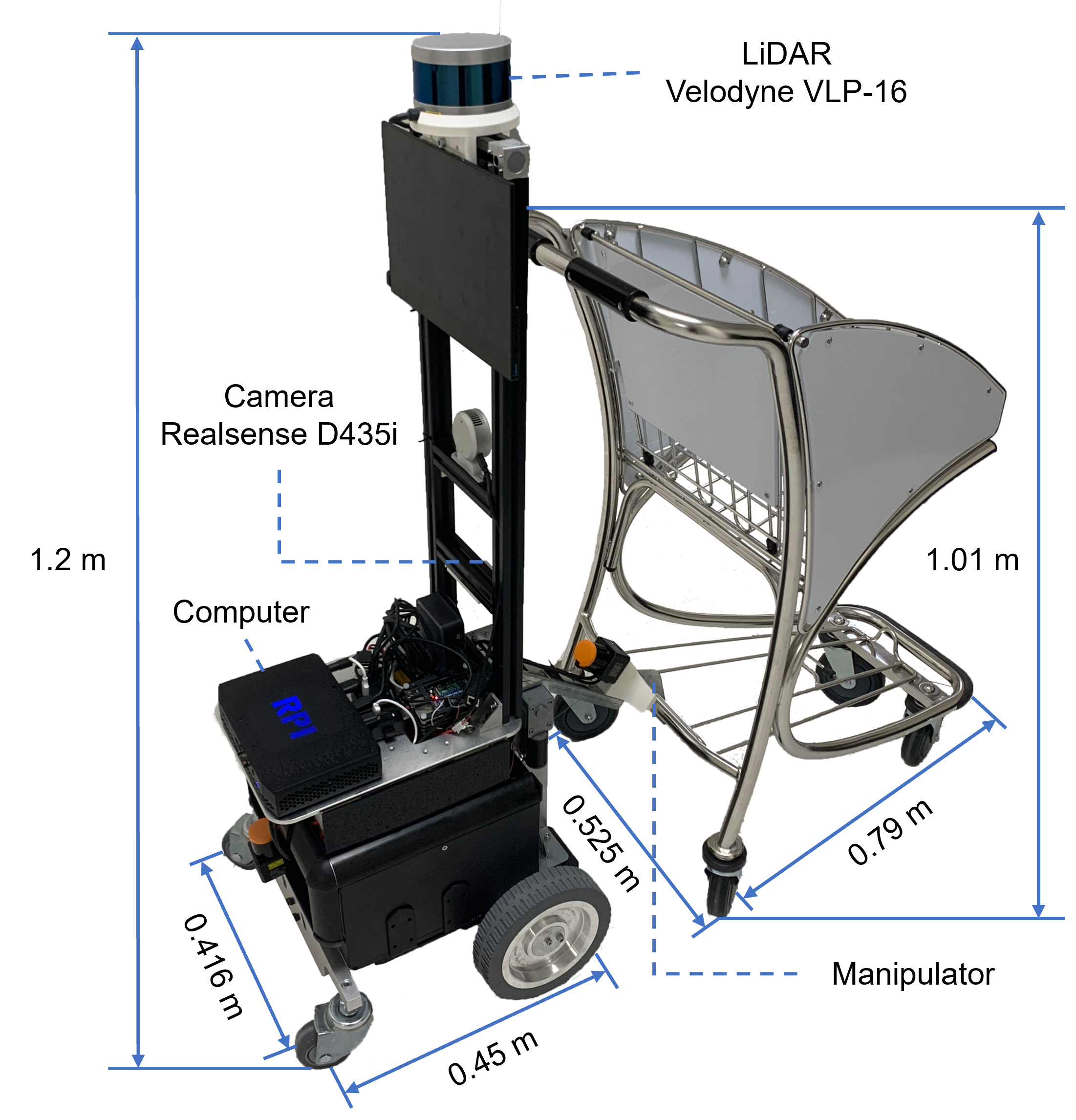}
    \caption{Experimental platform for real-world experiment.}
    \label{robot}
\end{figure}
As shown in Fig. \ref{robot}, this article employs a robot designed for luggage trolley collection. The robot measures 0.45 m $\times$ 0.416 m $\times$ 1.2 m, while the luggage trolley dimensions are 0.79 m $\times$ 0.525 m $\times$ 1.01 m. This robot features advanced sensors like LiDAR (Velodyne VLP-16) and a camera (Realsense D435i) powered by an onboard computer. It employs a specialized manipulator for the efficient collection of the luggage trolley. The algorithms integrate with the Robot Operating System (ROS) and function on the robot's onboard computer in real-time. This system is powered by an i7-1165G7 CPU and an NVIDIA GTX 2060 GPU.

\subsection{Experiment Results}
\subsubsection{Detection Results}
% To verify the effectiveness of our perception system, we conducted comparative experiments with various parts of the previous perception system. As shown in Tab \ref{trolley-detection}, we compared our method, Xiao's method and Xiao's improved method (mainly improving its ability to solve the luggage trolley pose under partial occlusion). It is worth noting that our method is retrained on the new data set, and Xiao's method and Xiao's improved method used its original training parameters.It is worth noting that our method is trained on our created dataset, and Xiao's method used its original training parameters.
To verify the effectiveness of HPPS detection, comparative experiments are conducted on various detection components of the previous perception system. We select categories in our dataset that match the detection capabilities of the Xiao's method. For instance, we chose the single idle luggage trolley during occlusion experiments, where Xiao's method struggles with multiple or occupied luggage trolleys. Despite this, our method consistently outperforms Xiao's in detection performance. Firstly, we compare our and Xiao's methods \cite{xiao} across different metrics for luggage trolley detection and classification, as shown in Tab. \ref{trolley-detection}. It evaluates the visibility of a single, idle luggage trolley at various visibility thresholds (\textgreater{80\%}, 40\% to 80\%, and \textless{40\%}), detection accuracy in scenarios with one to three luggage trolleys without occlusion, and luggage trolley classification accuracy when luggage trolleys are idle or occupied. It also assesses performance in complex situations where visibility is between 40\% to 80\% and only one of three luggage trolleys is idle. Xiao's method has varying success rates, is significantly lower in poor visibility conditions. In contrast, our method consistently achieves high accuracy, maintaining performance levels above 90\%, even in low visibility conditions, indicating a robust detection and classification even in complex situations. An important note is that Xiao's method fails to identify the states of the luggage trolley. However, this limitation has been effectively addressed with the help of our newly annotated dataset. To evaluate the accuracy of keypoint detection, we compare the HPPS with Xiao's method under the same luggage trolley detection and classification conditions, as detailed in Tab. \ref{keypoint-detection}. It presents the accuracy rates for keypoint detection across three visibility levels of luggage trolleys: above 80\%, between 40\% and 80\%, and below 40\%. We determine if a keypoint is successfully detected based on whether the sum of the average errors of its x and y coordinates does not exceed six pixels. Xiao's method shows a detection rate of 75\% for highly visible luggage trolleys, dropping to 33\% for moderate visibility and 9\% for low visibility. In contrast, our method significantly improves the detection rates to 89\% for high visibility, 70\% for moderate visibility, and 50\% for low visibility, clearly outperforming Xiao's approach across all categories.

\begin{table}[htb]
\centering
\caption{Comparison of Two Methods Regarding the Accuracy of Luggage Trolley Detection and Classification}
\label{trolley-detection}
\resizebox{0.48\textwidth}{!}{%
\renewcommand{\arraystretch}{1.8}
\huge % 
\begin{tabular}{cccc}
\toprule [3pt]
\multicolumn{2}{c}{\multirow{2}{*}{\textbf{Metric}}}                                                                                                                                                                      & \multicolumn{2}{c}{\textbf{Method}} \\ \cline{3-4} 
\multicolumn{2}{c}{}                                                                                                                                                                                                      & \textbf{Xiao’s Method} \cite{xiao}    & \textbf{Our Method}       \\ \hline
\multirow{3}{*}{\begin{tabular}[c]{@{}c@{}}Trolley Visibility \\      (Single Trolley, Idle)\end{tabular}}                                 & \textgreater{}80\%                                                           & 83\% (25/30)     & \textbf{100\%} (30/30)    \\
    & 40\% - 80\%                                                                  & 77\% (23/30)     & \textbf{97\%} (29/30)    \\
   & \textless 40\%                                                               & 20\% (6/30)      & \textbf{93\%} (28/30)    \\ \hline
\multirow{3}{*}{\begin{tabular}[c]{@{}c@{}}Trolley Count \\      (No Occlusion, Idle)\end{tabular}}                                        & One                                                                          & 93\% (28/30)     & \textbf{100\%} (30/30)    \\
            & Two                   & 92\% (55/60)     & \textbf{100\%} (60/60)    \\
                                   & Three                         & 81\% (73/90)     & \textbf{100\%} (90/90)    \\ \hline
\multirow{2}{*}{\begin{tabular}[c]{@{}c@{}}Trolley Classification \\      (Single Trolley, No Occlusion)\end{tabular}}                     & Idle                                                                         & —                & \textbf{100\%} (30/30)    \\
     & Occupied                       & —               & \textbf{100\%} (30/30)    \\ \hline
\begin{tabular}[c]{@{}c@{}}Complex Situations:\\ Poor Visibility\\ Multiple Trolley Counts\\ Multiple Trolley Classifications\end{tabular} & \begin{tabular}[c]{@{}c@{}}Recognition   of \\ the Idle Trolley\end{tabular} & —                & \textbf{93\%} (28/30)    \\ \bottomrule [3pt]
\end{tabular}%
}
\end{table}

% \usepackage{graphicx}
\begin{table}[htb]\tiny
\centering
\caption{Comparison of Two Methods Regarding the Keypoint Detection Accuracy}
\label{keypoint-detection}
\resizebox{\columnwidth}{!}{%
\renewcommand{\arraystretch}{1.5}
\begin{tabular}{ccc}
\toprule[1pt]
\textbf{Visibility}         & \textbf{Xiao’s Method} \cite{xiao}   & \textbf{Our Method}                    \\ \hline
\textgreater{}80\% & 75\% (113/150) & \textbf{89\%} (133/150) \\ 
40\% - 80\%        & 33\% (31/93)   & \textbf{70\%} (65/93)   \\ 
\textless 40\%     & 9\% (5/58)   & \textbf{50\%} (29/58)   \\ \bottomrule[1pt]
\end{tabular}%
}
\end{table}

\subsubsection{Localization Results}
\begin{figure}[htb]
    \centering
    \includegraphics[width=1\columnwidth]{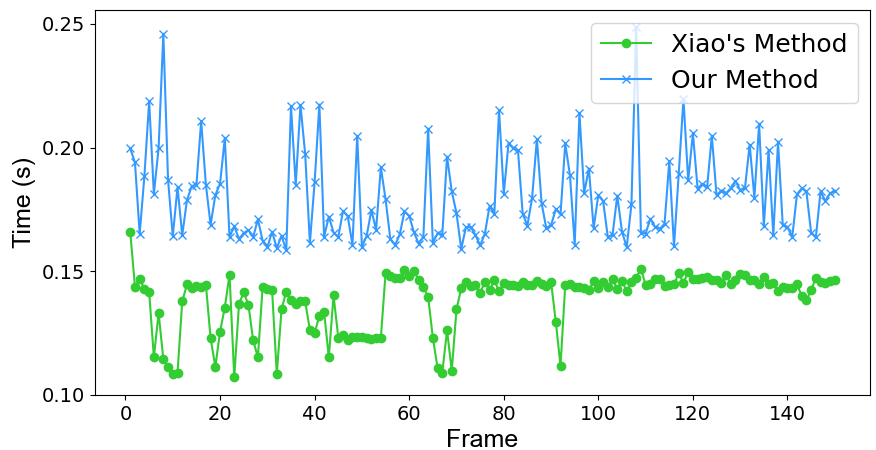}
    \caption{Running time comparison on the robot. The horizontal axis is the image frame, and the vertical axis is the running time.}
    \label{time}
\end{figure}
The two methods are implemented on the robot, and their operational real-time performance is assessed, as illustrated in Fig. \ref{time}. Over 150 image frames, the average execution time for our method is noted to be 0.180 ± 0.017 seconds, compared to Xiao's method at 0.138 ± 0.012 seconds. Although our approach shows slightly longer running times due to an additional network model for orientation estimation and a greater number of model parameters, it still meets the real-time requirements of robotic perception systems by delivering timely pose estimation.

\begin{figure}[htb]
    \centering
    \includegraphics[width=1\columnwidth]{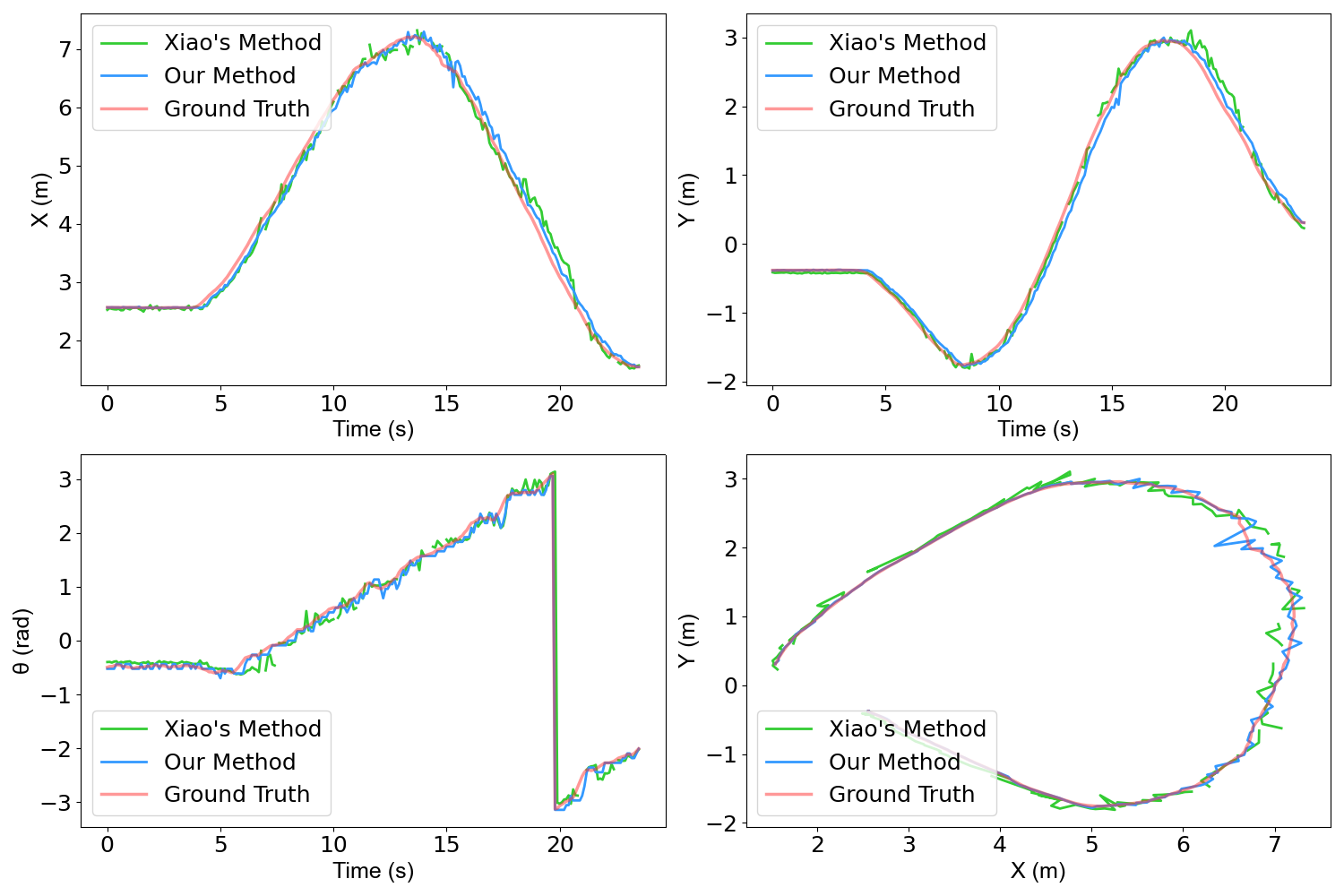}
    \caption{Comparison of ground truth, our method and Xiao's method for the poses of a moving luggage trolley with no occlusion. The four subfigures show the $x$ coordinate, $y$ coordinate, orientation ($\theta$), and moving trajectory, respectively.}
    \label{no_ob}
\end{figure}

\begin{figure}[htb]
    \centering
    \includegraphics[width=1\columnwidth]{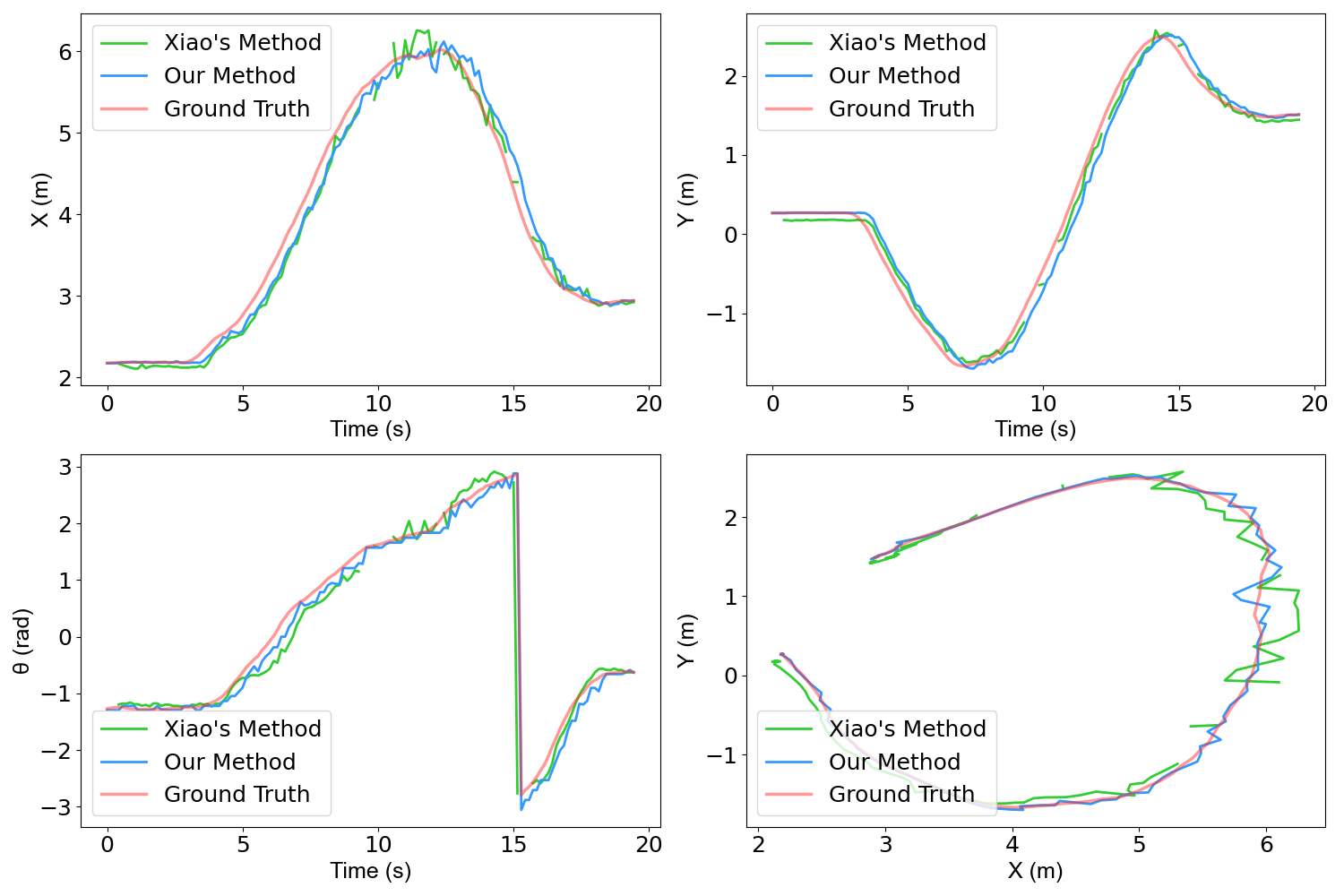}
    \caption{Comparison of ground truth, our method and Xiao's method for the poses of a moving luggage trolley with static obstacle occlusion. The four subfigures show the $x$ coordinate, $y$ coordinate, orientation ($\theta$), and moving trajectory, respectively.}
    \label{simple_ob}
\end{figure}

\begin{figure}[htb]
    \centering
    \includegraphics[width=1\columnwidth]{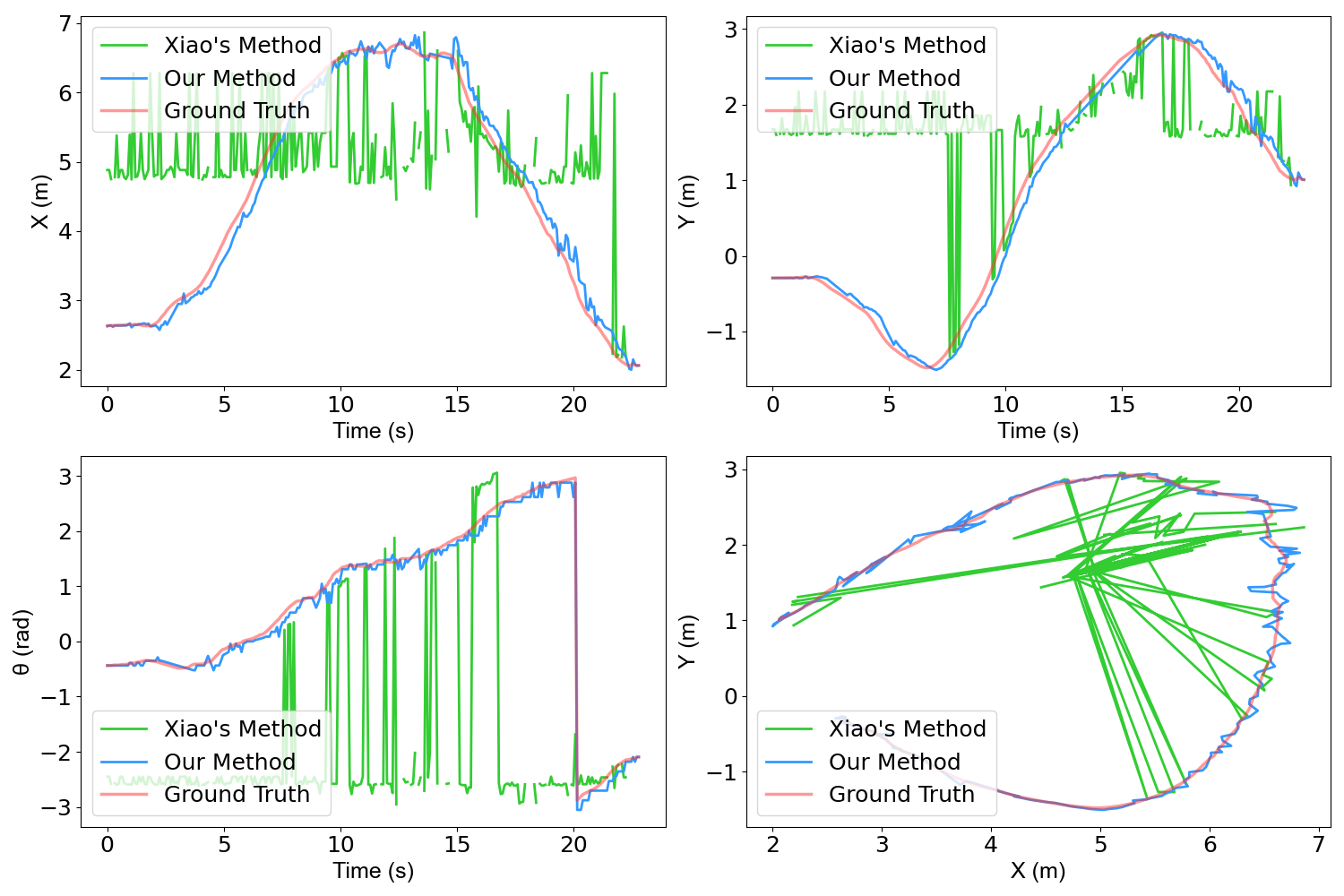}
    \caption{Comparison of ground truth, our method and Xiao's method for the poses of a moving luggage trolley with occupied trolley occlusion. The four subfigures show the $x$ coordinate, $y$ coordinate, orientation ($\theta$), and moving trajectory, respectively.}
    \label{complex_ob}
\end{figure}

\begin{table*}[htb]
\centering
\caption{The Mean and Standard Deviation of $x$, $y$, and $\theta$ for Two Methods Under Three Different Occlusion Conditions}
\label{mean-std}
\resizebox{2\columnwidth}{!}{%
\renewcommand{\arraystretch}{1.7}
\begin{tabular}{ccccccc}
\toprule[1pt]
\multirow{2}{*}{\textbf{Occlusion Conditions}} & \multicolumn{3}{c}{\textbf{Xiao's Method} \cite{xiao}}           & \multicolumn{3}{c}{\textbf{Our Method}}             \\ \cline{2-7} 
                                                 & $\bm{x}$     & $\bm{y}$       & $\bm{\theta}$    & $\bm{x}$      & $\bm{y}$      & $\bm{\theta}$    \\ \hline
No Occlusion                                     & 0.1199 ± 0.1223 & 0.0766 ± 0.0882  & 0.1285 ± 0.4217 & 0.1136 ± 0.0766 & 0.0933 ± 0.0703  & 0.0710 ± 0.0653  \\ 
Occluded   by Static Obstacles                   & 0.1428 ± 0.1024 & 0.0961 ± 0.0485  & 0.1978 ± 0.5082 & 0.1320 ± 0.1132 & 0.1312 ± 0.1091 & 0.1228 ± 0.1066 \\ 
Occluded by an Occupied Luggage Trolley         & 1.2550 ± 0.9181 & 1.3454 ±  1.1319 & 2.6815 ± 1.7123 & 0.1496 ± 0.1067 & 0.1209 ± 0.0841 & 0.0848 ± 0.0719 \\ \bottomrule[1pt]
\end{tabular}%
}
\end{table*}

HPPS and Xiao's method are tested on a robot to evaluate their ability to detect the pose of a moving luggage trolley. The 3D poses detected by these two methods are compared with ground truth measured by a motion capture system. Three conditions are tested: no occlusion, static obstacles occlusion, and luggage trolley occlusion, as illustrated in Fig. \ref{no_ob}, \ref{simple_ob}, and \ref{complex_ob}, respectively. It is important to note that the version of Xiao's method used in this experiment has been modified. Initially, Xiao's method requires identifying six keypoints to determine the pose of the luggage trolley. However, the EPNP calculation needs only four or more keypoints. Thus, Xiao's method has been optimized here to ensure pose estimation with four or more keypoints. 

In scenarios without occlusion (refer to Fig. \ref{no_ob}), both methods effectively capture the pose of the luggage trolley. However, self-occlusion during the luggage trolley's movement results in incomplete trajectory tracking by Xiao's method, while our method consistently achieves complete recognition. With static obstacles present (refer to Fig. \ref{simple_ob}), the limitations of Xiao's method become more apparent, particularly in sections occluded by obstacles where the luggage trolley's pose is frequently lost. Conversely, our method maintains complete acquisition of the luggage trolley. In situations where the occupied luggage trolley is an occlusion (refer to Fig. \ref{complex_ob}), Xiao's method struggles with distinguishing the trolley's states, often failing to capture the pose of the target luggage trolley.
In contrast, our method effectively and accurately detects and locates the idle luggage trolley. Furthermore, these three experimental results illustrate that as the detection distance surpasses 5 meters, oscillations in the pose obtained by HPPS increase, leading to a decrease in locating accuracy. This decrease in accuracy originates from the precision of keypoints's image coordinates. As the proportion of the locating object within the image becomes minimal, even minor errors in image coordinates lead to significant inaccuracies in locating. In contrast, at closer detection distances, the locating exhibits greater accuracy and smoothness. Thus, a progressive perception strategy has been integrated into this system. Following the acquisition of the luggage trolley' initial pose, the system continuously updates the luggage trolley's pose with effective observation information (refer to Sec. \ref{filter}), thereby reducing errors associated with long-distance locating.

Tab. \ref{mean-std} provides a more detailed comparison of the mean and standard deviations for the proposed HPPS and Xiao's method for measurements of horizontal distance ($x$), vertical distance ($y$), and angle ($\theta$). The table shows that the proposed method consistently maintains the mean and standard deviation of the locating error, regardless of occlusion. In contrast, Xiao's method shows increased locating errors under occlusion conditions, particularly in scenarios with multiple luggage trolleys. Although the mean and standard deviation values of Xiao's method under static obstacles occlusion seem similar to ours, this similarity is because Xiao's method does have many occlusion data, leading to incomplete statistical results. Nonetheless, our method equals or exceeds Xiao's method in performance, demonstrating greater resilience to occlusion and improved locating robustness.

\subsubsection{Robot Trial Results}
\begin{figure*}[t]
    \centering
    \includegraphics[width=2\columnwidth]{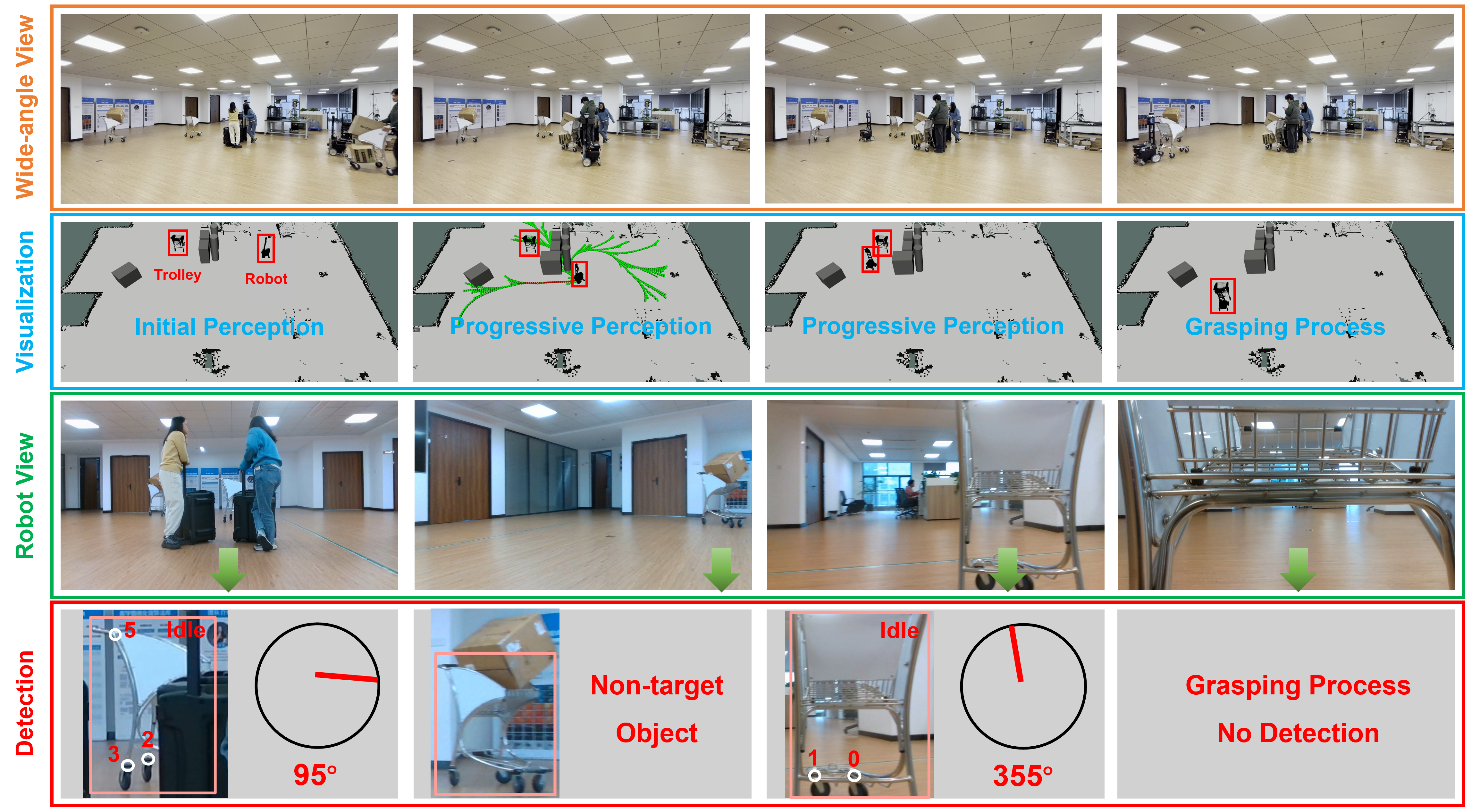}
    \caption{Snapshots from the real-world experimental process. The orange, blue, and green boxes signify three different views. The big red box represents the detection of the luggage trolley's classification, keypoints, and orientation from the robot's view. The small red box represents the robot's and luggage trolley's position at different moments in the visualization.}
    \label{real-world}
\end{figure*}
As shown in Fig. \ref{real-world}, a real-world experiment is constructed to verify the proposed perception system, HPPS. To simulate the airport environment, two pedestrians with luggage are arranged to stop and chat, and another pedestrian pushes a luggage trolley. Through HPPS, the robot initially detects an idle luggage trolley under partial occlusion. It then locates the luggage trolley and navigates towards it while avoiding pedestrians and obstacles. HPPS filters out interference from occupied luggage trolleys during navigation and refines the target's pose as the robot moves closer. Finally, the robot grasps the luggage trolley and stops detection during transportation.
% \section{Discussion}
% Discussion.
\section{Conclusions and Future Work}\label{future}
This article introduces the HPPS, which detects the position and orientation of luggage trolleys separately and gradually updates the luggage trolley's pose during navigation. This innovative hierarchical processing structure simplifies the labeling needs for datasets and enhances practicality and scalability. Additionally, the system's progressive perception improves robustness and accuracy for locating luggage trolleys. The HPPS robustly completes the localization task, even in cases of partial occlusion. We have tested the system's accuracy and robustness in experiments and demonstrated its capabilities on actual trolley collection tasks in complex environments. Thus, the HPPS enhances the robot's ability to detect and localize luggage trolleys, providing a foundation for the subsequent operation and execution of the robotic autonomous luggage trolley collection system.

In future work, we intend to explore the potential of multi-robot cooperation to enhance environment perception and further advance our research of multi-robot motion planning.

\section*{Conflict of Interest}
Jiankun Wang is an Associate Editor of IEEE Transactions on Intelligent Vehicles.

\bibliographystyle{IEEEtran} % We choose the "plain" reference style
\bibliography{refs} % Entries are in the refs.bib file

% \afterpage{
% \vspace{4 mm}
\begin{IEEEbiography}
[{\includegraphics[width=1in,height=1.25in,clip,keepaspectratio]{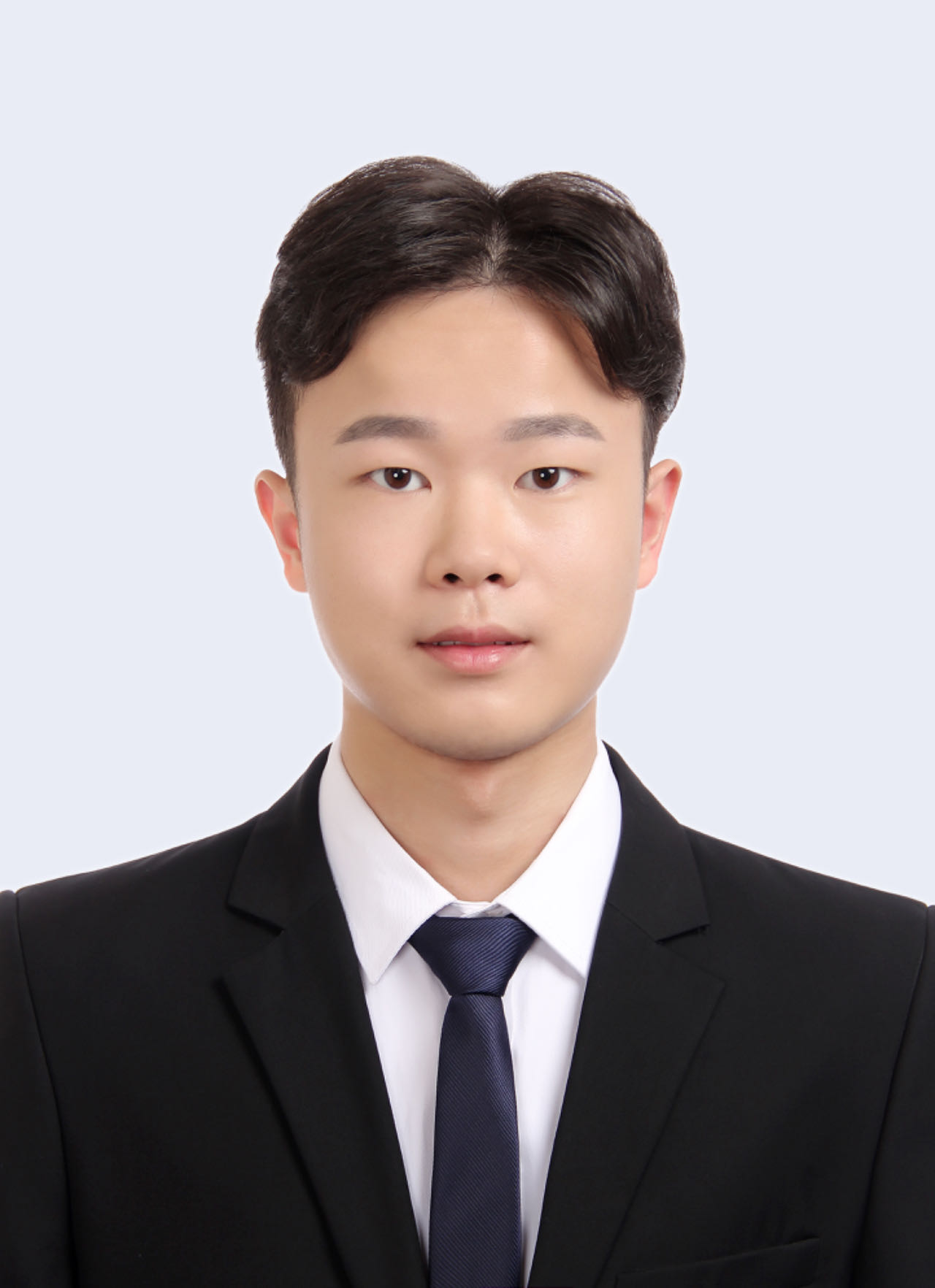}}] 
{Zhirui Sun} received the B.E. degree in information engineering from the Department of Electronic and Electrical Engineering, Southern University of Science and Technology, Shenzhen, China, in 2019. He is currently pursuing the Ph.D. degree with the Department of Electronic and Electrical Engineering, Southern University of Science and Technology, Shenzhen, China. His research interests include robot perception and motion planning.
\end{IEEEbiography}
\vspace{20 mm}

\begin{IEEEbiography}
[{\includegraphics[width=1in,height=1.25in,clip,keepaspectratio]{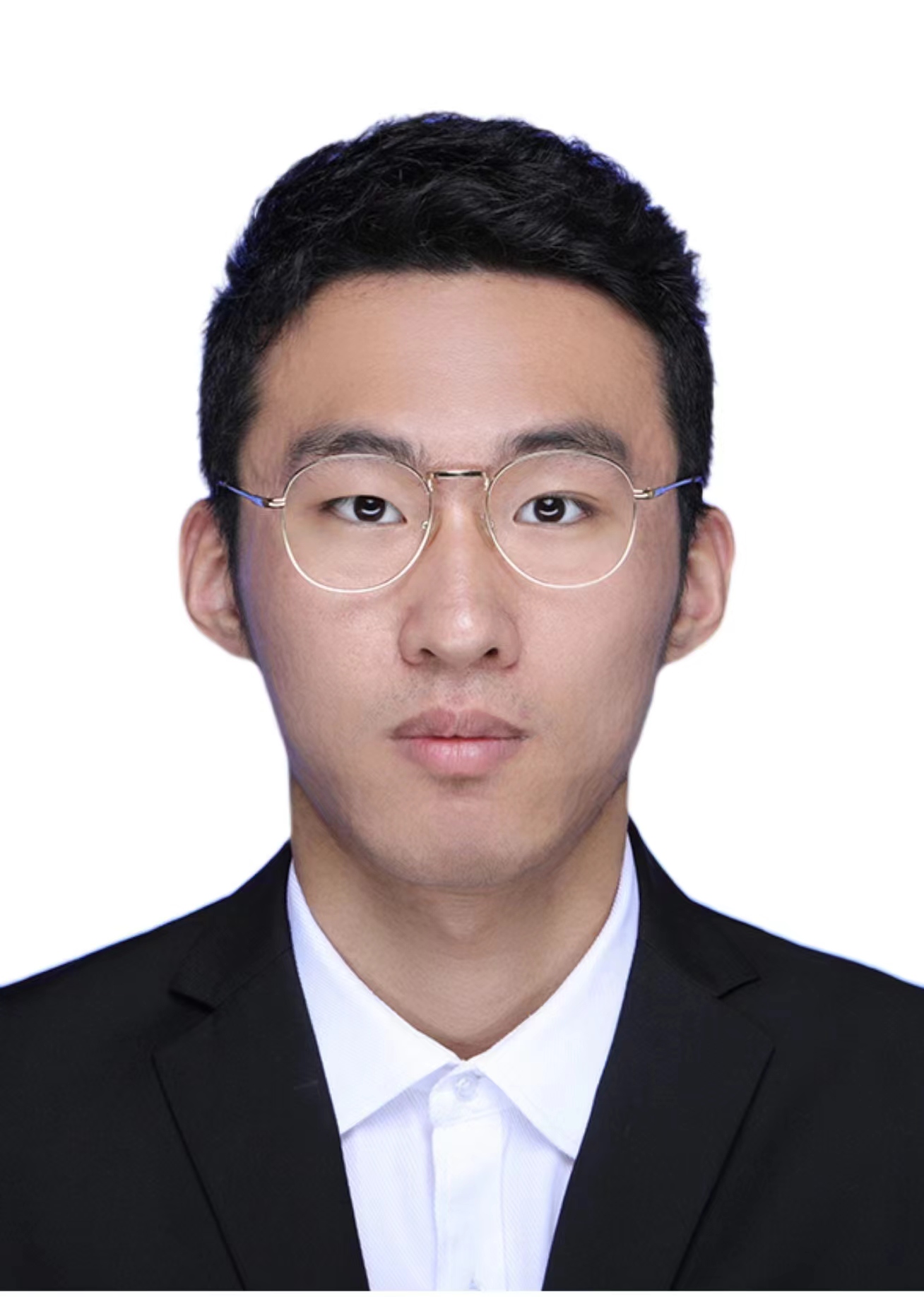}}] 
{Zhe Zhang} received his B.E. degree in Electrical Engineering from Beijing Jiaotong University, Beijing, China, in 2022. He is currently pursuing the M.S. degree with the Department of Electronic and Electrical Engineering, Southern University of Science and Technology, Shenzhen, China. His research interests include human-robot interaction and motion planning.
\end{IEEEbiography}

\vspace{20 mm}
\begin{IEEEbiography}
[{\includegraphics[width=1in,height=1.25in,clip,keepaspectratio]{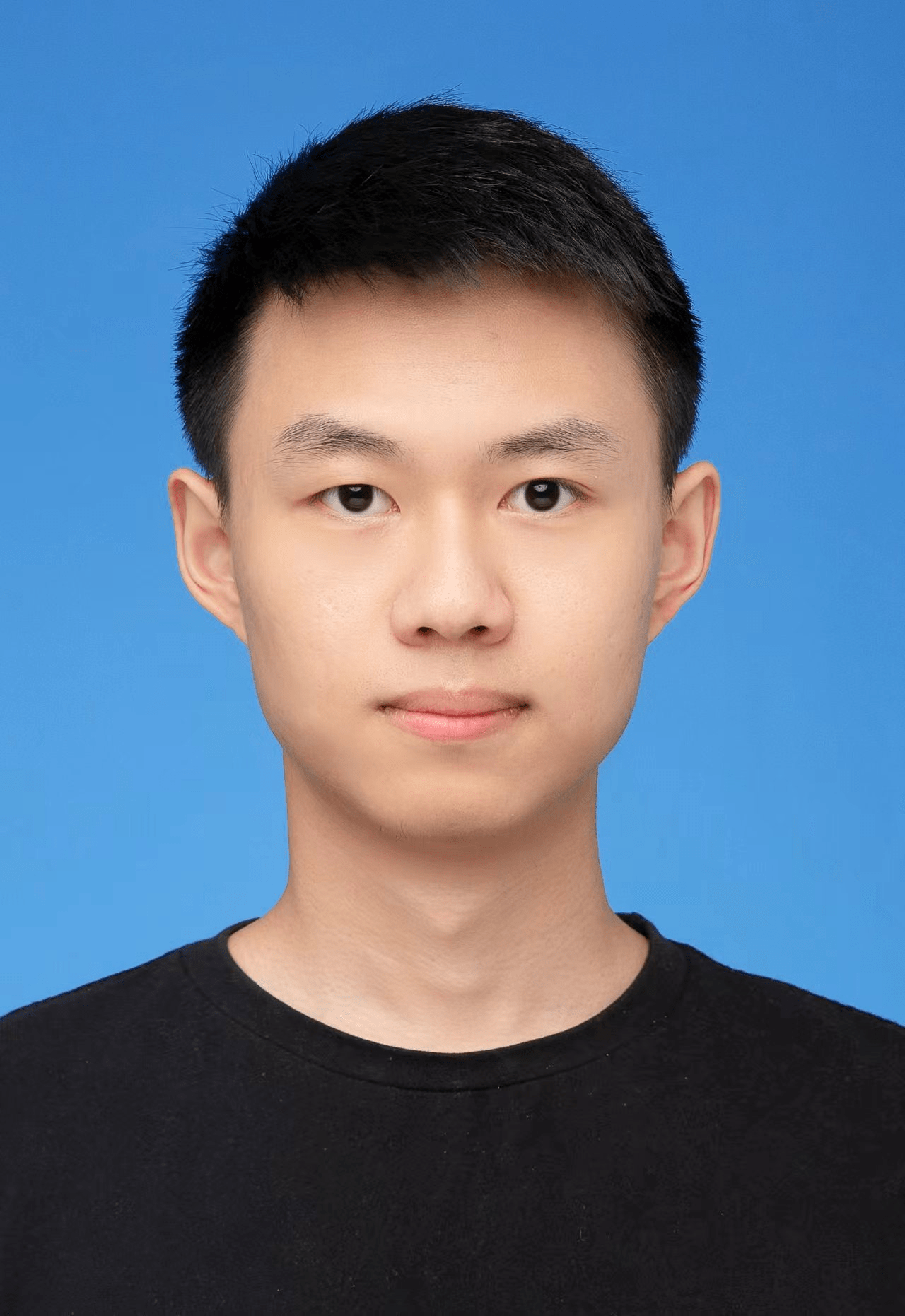}}] 
{Jieting Zhao} received a B.E. degree in information engineering from the China University of Mining \& Technology, Beijing, China, in 2021. He is pursuing the M.S. degree with the Department of Electrical and Electronic Engineering, Southern University of Science and Technology, Shenzhen, China. His research interests involve human-robot interaction, person-following robot, and robot vision.
\end{IEEEbiography}

\vspace{20 mm}
\begin{IEEEbiography}
[{\includegraphics[width=1in,height=1.25in,clip,keepaspectratio]{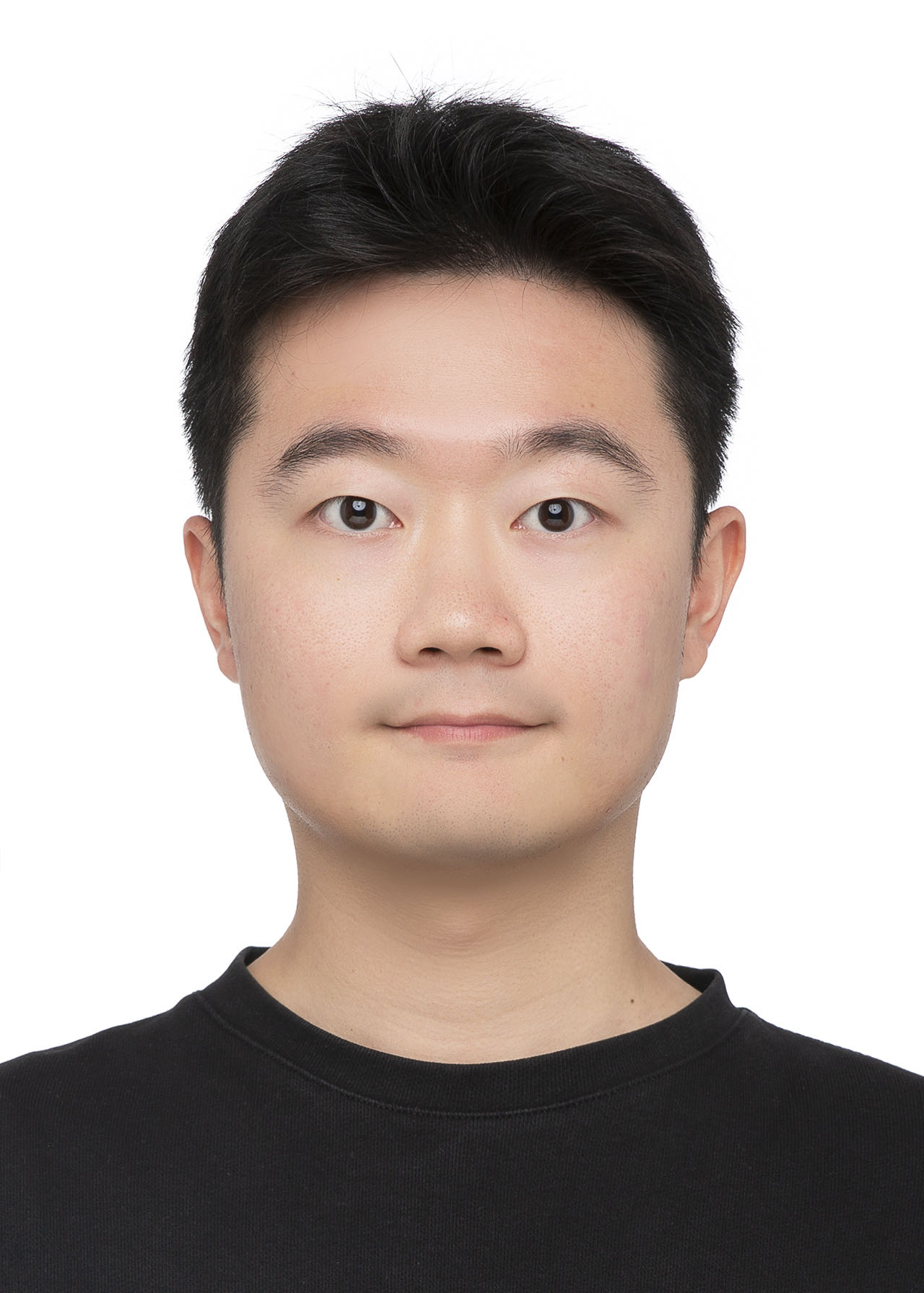}}] 
{Hanjing Ye} received the B.E. degree and M.S. degree from Guangdong University of Technology,  Guangzhou, China, in 2019 and 2021, respectively. He is currently a Ph.D. candidate at Department of Electronic and Electrical Engineering, Southern University of Science and Technology in Shenzhen, China. His research interests in human-robot interaction and mobile robot navigation.
\end{IEEEbiography}
\vspace{20 mm}

\begin{IEEEbiography}
[{\includegraphics[width=1in,height=1.25in,clip,keepaspectratio]{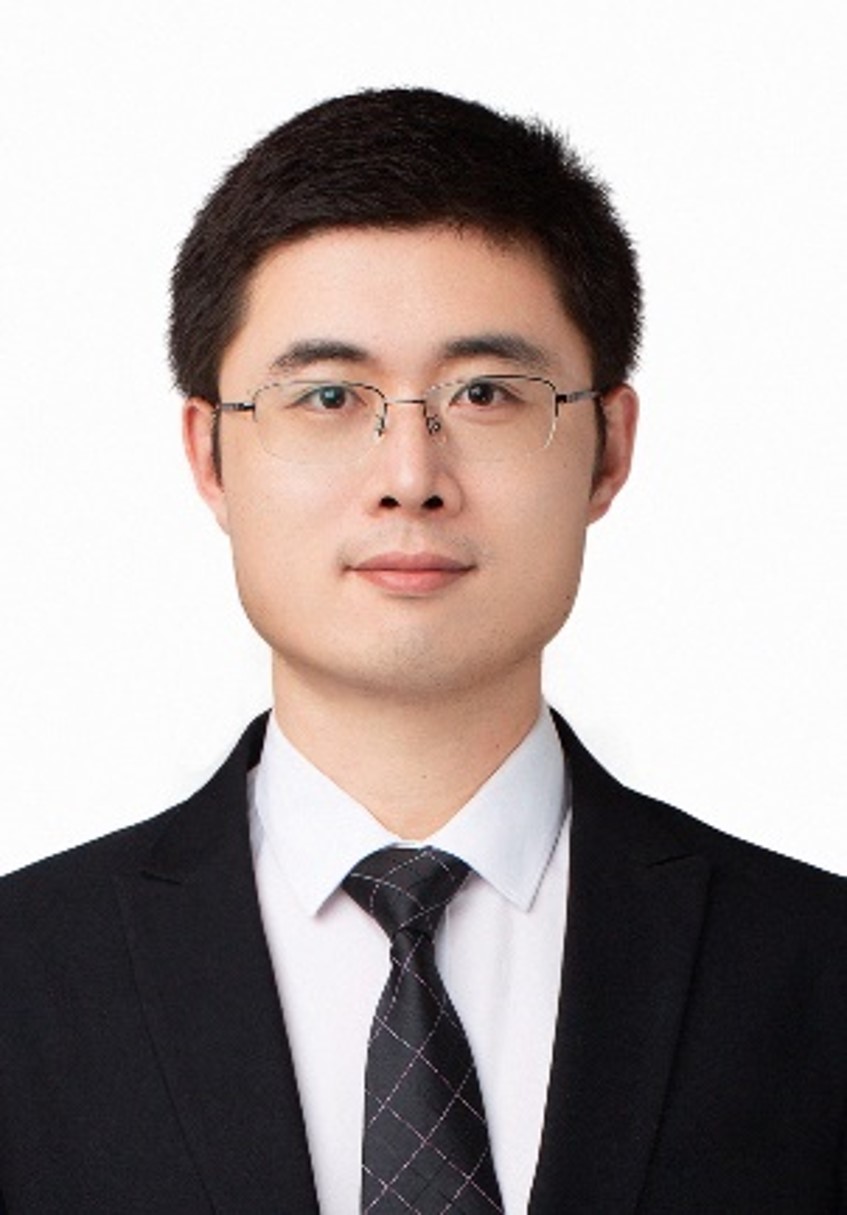}}] 
{Jiankun Wang} (Senior Member, IEEE) received the B.E. degree in automation from Shandong University, Jinan, China, in 2015, and the Ph.D. degree from the Department of Electronic Engineering, The Chinese University of Hong Kong, Hong Kong, in 2019.

During his Ph.D. degree, he spent six months with Stanford University, Stanford, CA, USA, as a Visiting Student Scholar, supervised by Prof. Oussama Khatib. He is currently an Assistant Professor with the Department of Electronic and Electrical Engineering, Southern University of Science and Technology, Shenzhen, China. His current research interests include motion planning and control, human–robot interaction, and machine learning in robotics.

Currently, he serves as the associate editor of IEEE Transactions on Automation Science and Engineering, IEEE Transactions on Intelligent Vehicles, IEEE Robotics and Automation Letters, International Journal of Robotics and Automation, and Biomimetic Intelligence and Robotics.
\end{IEEEbiography}

\end{document}